\documentclass{article}
\usepackage[margin=3cm]{geometry}
\usepackage[utf8]{inputenc}
\usepackage{amsmath}
\usepackage{amsthm}
\usepackage{amsfonts}
\usepackage{float}
\usepackage{graphicx}
\usepackage{bm}
\usepackage{amssymb}
\usepackage{bbm}
\usepackage{color}
\usepackage{natbib}
\usepackage{algorithm}
\usepackage{algorithmic}
\usepackage{mathtools}

\usepackage{ dsfont }

\newtheorem{theorem}{Theorem}[section]
\newtheorem{lemma}[theorem]{Lemma}
\newtheorem{definition}[theorem]{Definition}

\newtheorem{assumption}[theorem]{Assumption}

\newcommand{\R}{\mathbb{R}}
\newcommand{\D}{\mathcal{D}}
\newcommand{\X}{\mathcal{X}}

\newcommand{\mC}{\mathcal{C}}

\newcommand{\mS}{\mathcal{S}}
\newcommand{\mT}{\mathcal{T}}

\newcommand{\mA}{\mathcal{A}}

\newcommand{\mG}{\mathcal{G}}

\newcommand{\Split}{\mathrm{Split}}
\newcommand{\Merge}{\mathrm{Merge}}

\newcommand{\multiA}{\mathrm{ExpUpdate}}

\newcommand{\y}{\mathbf{y}}
\newcommand{\x}{\mathbf{x}}

\newcommand{\s}{\mathbf{S}}

\newcommand{\logf}[2]{\ensuremath{\log\left(\frac{#1}{#2}\right)}}
\newcommand{\rgta}{\rightarrow}

\newcommand{\zo}{\ensuremath{\{0,1\}}}

\newcommand{\eat}[1]{}
\newcommand{\KL}[2]{D\left( #1 \| #2 \right)}

\newcommand{\dtv}{\mathrm{d_{TV}}}
\newcommand{\qme}{\ensuremath{Q^{\mathrm{ME}}}}
\newcommand{\wme}{\ensuremath{w^{\mathrm{ME}}}}

\newcommand{\norm}[1]{\left\lVert#1\right\rVert}

\newcommand{\infnorm}[1]{\left\lVert#1\right\rVert_\infty}

\newcommand{\Macc}{Multi-accuracy}
\newcommand{\macc}{multi-accuracy}
\newcommand{\mact}{multi-accurate}
\newcommand{\Mcab}{Multi-calibration}
\newcommand{\mcab}{multi-calibration}
\newcommand{\mcbd}{multi-calibrated}
\newcommand{\maxent}{MaxEnt}

\renewcommand{\varepsilon}{\epsilon}

\newcommand{\lb}{\ensuremath{\left(}}
\newcommand{\rb}{\ensuremath{\right)}}
\newcommand{\abs}[1]{\ensuremath \Bigl\lvert #1 \Bigr\rvert}

\DeclareMathOperator*{\C}{\mathbb{C}}

\DeclareMathOperator*{\E}{\mathbb{E}}

\newcommand{\VS}[1]{{\color{blue}[Vatsal: #1]}}
\newcommand{\UW}[1]{{\color{green}[Udi: #1]}}

\begin{document}
\newcommand{\theTitle}{Multicalibrated Partitions for Importance Weights}

\author{Parikshit Gopalan\footnote{Email: \texttt{pgopalan@vmware.com}}\\
VMware Research
\and Omer Reingold\footnote{Most of the work performed while visiting VMware Research. Research supported in part by
NSF Award IIS-1908774. Email: \texttt{reingold@stanford.edu}}\\
Stanford University
\and Vatsal Sharan\footnote{Most of the work performed while visiting VMware Research. Email: \texttt{vsharan@mit.edu}}\\
MIT
\and Udi Wieder\footnote{Email: \texttt{uwieder@vmware.com}}\\
VMware Research
}

\title{\theTitle}
\date{}

\clearpage
\maketitle\begin{abstract}
The ratio between the probability that two distributions $R$ and $P$ give to points $x$ are known as importance weights or propensity scores and play a fundamental role in many different fields, most notably, statistics and machine learning. Among its applications, importance weights are central to domain adaptation, anomaly detection, and estimations of various divergences such as the KL divergence and Renyi divergences between $R$ and $P$, which in turn have numerous applications. We consider the common setting where $R$ and $P$ are only given through samples from each distribution. The vast literature on estimating importance weights is either heuristic, or makes strong assumptions about $R$ and $P$ or on the importance weights themselves. Indeed, relying on cryptographic assumptions, we show the impossibility of efficiently computing pointwise accurate importance weights.

In this paper, we explore a computational perspective to the estimation of importance weights, which factors in the limitations and possibilities obtainable with bounded computational resources. We significantly strengthen previous work that use the MaxEntropy approach, that define the importance weights based on a distribution $Q$ closest to $P$, that looks the same as $R$ on every set $C \in \mathcal{C}$, where $\mathcal{C}$ may be a huge collection of sets. We show that the MaxEntropy approach may fail to assign high average scores to sets $C \in \mathcal{C}$, even when the average of ground truth weights for the set is evidently large. We similarly show that it may overestimate the average scores to sets $C \in \mathcal{C}$. We therefore formulate Sandwiching bounds as a notion of set-wise accuracy for importance weights. We study these bounds to show that they capture natural completeness and soundness requirements from the weights and are appealing  from the point of view of accuracy and fairness in heterogeneous populations. We present an efficient algorithm that under standard learnability assumptions computes weights which satisfy these bounds.

Our techniques rely on a new notion of multicalibrated partitions of the domain of the distributions. While being a relatively small collection of disjoint sets, such partitions reflect, in a well-defined sense, the complexity of the much larger collection of arbitrarily intersecting sets in $\mathcal{C}$. Stratifying the domain based on multicalibrated partitions implies our computational objectives for importance weights and appear to be useful objects in their own right.
    
\end{abstract}
\thispagestyle{empty}
\newpage
\tableofcontents
\newpage
\setcounter{page}{1}

\eat{
\maketitle

}
\section{Introduction}\label{sec:intro}

Consider a scenario where we are observing samples drawn from a distribution $R$ over some domain $\X$. The domain $\X$ might represent demographic information of people, whereas $R$ might be the distribution of people who voted, who applied to college or were afflicted with the flu virus. We have some knowledge about the distribution, perhaps from previous experience, which is captured by a prior distribution $P$. While $P$ could be a good baseline, it is not necessarily a good model for $R$. How can we modify $P$ to better represent $R$? The statistical technique of \emph{importance weighting}, also called density ratio estimation, provides an approach to this problem. For each $x \in \X$, let us define its importance weight under $R$ to be $w^*(x) = R(x)/P(x)$. This defines a function $w^*: \X \rgta \R^+$.

The problem of estimating these importance weights has been studied by several communities (often under different monikers) and is central to a wide variety of applications. In machine learning, it arises both in unsupervised problems such as anomaly detection and supervised settings such as in domain adaptation. For anomaly detection, the objective is to find points which have a small probability under a prior $P$ but high probability under the observed distribution $R$; essentially the points with high importance weights \citep{scholkopf2001estimating,hodge2004survey,hido2008inlier,smola2009relative}. In domain adaptation (also known as covariate shift), we get labelled training data from a distribution $P$ but are interested in  good prediction accuracy under a different test distribution $R$ for which we only observe unlabelled data. The importance weights of $R$ under $P$ can be used to reweigh the loss function to correct for the distributional shift \citep{shimodaira2000improving,zadrozny2004learning,sugiyama2005input,bickel2007discriminative,cortes2010learning}.

In information theory,  importance weight estimation (where it is sometimes known as Radon-Nikodym derivative estimation), is applied to estimate various divergences such as the KL divergence and Renyi divergences between the distributions $P$ and $R$ \citep{nguyen2007nonparametric,nguyen2010estimating,yamada2013relative,wang2005divergence,wang2009divergence}. These divergence measures themselves find several applications, such as two-sample tests for distinguishing between two distributions \citep{wornowizki2016two} and for independence testing \citep{suzuki2008approximating}.

In econometrics and statistics, importance weights (under the guise of propensity scores) play a major role in the theory of causal inference from observational data. Propensity scores denote the probability of an individual being selected for a treatment---this is just a scaling of the importance weights of the distribution $R$ of the individuals selected for the treatment, under the prior $P$ of the entire population. These scores are widely used to correct for bias introduced by confounding variables which influence the probability of getting the treatment itself (see survey by \cite{austin2011introduction}). Starting with \cite{rosenbaum1983} there is a rich body of work focused on estimating propensity scores \citep{hirano2003,imai2014,stuart2011}. 

\eat{To provide some more grounding for our discussion, we give an example in the context of anomaly detection. }
Let us examine the role of importance weights in anomaly detection more closely. 
In the semi-supervised setting \citep{chandola2009anomaly}, we are given samples from $R$ and wish to identify anomalous points or regions. Our prior knowledge is encoded by a distribution $P$ (for which we might even have a closed form). A simple approach might be to identify points that are unlikely under $P$ as anomalies, after all, seeing an unlikely point is indeed more surprising than seeing a likely point. However this is not a good approach for defining anomalies. For example, suppose $P$ is Gaussian, and $R$ has far more points close to the mean than $P$ predicts, as was the case with Mendel’s experiments in genetics \citep{fisher1936has}. Points in the tail of the distribution $P$ would be considered anomalies, even though the number of such points is lower than expected, while the excess density near the mean would be missed. In other words, we do expect low-probability points to appear in our sample of $R$ and they should only be considered anomalous if they appear in $R$ with higher probability than in $P$. The importance weights of $R$ under $P$ do exactly that and are therefore more suitable as a measure of anomaly. In the example above, they correctly flag the excess density near the mean as being anomalous.

\subsection{Computational Limitations}

Given the dramatic impact of importance weights in so many research fields and applications, a natural question is how do we compute them, and what guarantees can one hope to achieve. We consider the setting where we only have access to $R$ and $P$ through random samples drawn from the distributions. Not only is this the most restrictive setting, it is the realistic assumption in most applications, for example when we want to determine how the spending patterns of customers this month differ from twelve months ago. 

Computing the true weights $w^*(x) = R(x)/P(x)$ for each point based on samples from $R$ and $P$ is impossible in general (we will shortly justify it under standard cryptographic assumptions). Much of the literature, in different communities, makes strong assumptions on $R$ and $P$, or on the ratio function $w^*(x) = R(x)/P(x)$ which make learning the weights possible. It is not obvious that such assumptions are justified in the real-world applications that motivate these works. Even if the calculated weights are approximately correct on average, it is less obvious why they would be correct on various sub-populations. This raises concerns of algorithmic discrimination: imagine for example image processing software used by a security firm that identifies movement by individuals of a particular ethnicity as anomalous with unfairly high probabilities. 

{\bf A cryptographic barrier: } Barring any distributional assumptions, {\em can we non-trivially approximate the importance weights in polynomial time?} Under standard cryptographic assumptions, the answer is that {\em we cannot}: Let $R^0 \equiv P$ be the uniform distribution on $\zo^d$. Let $R^1$ be the output of a cryptographically secure pseudorandom generator \citep{goldreich2000candidate} on a uniform seed. Every $x$ has importance weight $1$ under $R^0$. On the other hand, with probability $1$, an $x$ sampled from $R^1$ will have importance weights exponential in $d$. If an algorithm only gets access to random samples from $R$, it cannot distinguish $R = R^0$  from $R = R^1$ and hence cannot accurately determine whether every observation is hugely anomalous or perfectly normal---despite the huge gap in the weights.  We borrow the terminology of anomaly detection, but this observation is relevant much more broadly. We observe that in this example, the case $R^0 =  P$ is as simple as it gets and that the example could be generalized to other distributions (as long as they have large enough entropy).

This example shows that for arbitrary distributions $P$ and $R$, {\bf accuracy for point-wise scores is impossible}. It suggests that for provable guarantees, we need to look for {\bf set-wise accuracy guarantees}, and restrict our attention to a collection of \emph{nice} sets, whose complexity we constrain so as to exclude bad examples such as the support of a pseudorandom generator. We will model this by a collection of sets ${\mC}\subseteq 2^\X$, where $\X$ is the domain of the distributions $R$ and $P$, which will define the statistical tests that our weights must pass. Membership in every set $C \in \mC$ has to be easy to test (we will subsequently add some learnability assumptions). $\mC$ may represent an algorithm's computational resources in the sense that the algorithm really only uses samples from a distribution $Q$ to compute $Q(C) = \sum_{x \in C}Q(x)$ for $C\in \mC$. This would correspond to a statistical query algorithm that queries sets in $\mC$. In the fairness context, we would like $\mC$ to capture the protected subgroups and any other sub-population for which we want guarantees on our weights.

\subsection{Sandwiching Bounds}

In order to formulate the set-wise accuracy guarantees we wish to achieve, let us consider the setting of anomaly detection. Consider a researcher analyzing health data, where each point represents a patient. $R$ represents the distribution of data they actually see, whereas $P$ represents a prior. $\mC$ is a (possibly huge) collection of medical conditions that can be diagnosed from the data. 
A learning algorithm has assigned  importance weights $w:\X \rgta \R$ to the points. We articulate two desirable properties of the weights.
\begin{enumerate}
    \item Let $C \in \mC$ be a medical condition such  that $R(C)/P(C) > 10$, so that the prevalence of $C$ in the real world is $10$ times higher than was expected in the prior.  The researcher would like a random $R$-sample from $C$ to be assigned large weight by $w$, ideally at least $10$. If not, $w$ might not alert them to the increased prevalence of $C$. 
    \item Let $C' \in \mC$ be a medical condition such that a random $R$-sample from $C'$ is assigned an average weight of $10$ by $w$. The researcher would like this to imply that $C'$ is truly important under $R$ in some precise sense. If not, having large weights $w$ in expectation for $C'$ is not a reliable signal of importance under $R$ and might be a false alarm. 
\end{enumerate}
In analogy to proof systems, these conditions ask for {\bf completeness} and {\bf soundness} of the importance weights respectively. Completeness requires that if a set $C$ is important under $R$, then it receives large weights $w$ on average. Soundness requires that if the average weight under $R$ assigned to a set $C'$ is large, this indicates that the set is important.

 We now rigorously formulate these intuitive requirements. For a distribution $R$ and a set $C$, let $R|_C$ denote the distribution $R$ conditioned on $C$. We would like the following {\bf Sandwiching bounds} to hold for every $C \in \mC$:
\begin{align}
\label{eq:sandwich0}
    \E_{\x \sim P|_C}[w^*(\x)] \leq \E_{\x \sim R|_C}[w(\x)]  \leq \E_{\x \sim R|_C}[w^*(\x)].
\end{align}
The quantity in the middle is one that we can compute from random samples of $R$, given $w$. We want it to be sandwiched between the expectations of the ground-truth scores $w^*$ under $P|_C$ and $R|_C$ (note that we don't have $w^*$ explicitly).  

Let us see why sandwiching bounds indeed capture the aforementioned requirements. For the lower bound, the expectation on the left can be written as
\begin{align*}
    \E_{\x \sim P|_C}[w^*(\x)] = \frac{R(C)}{P(C)}.
\end{align*}
Hence this inequality captures condition (1). For the upper bound, we want our weights to be conservative, they should not exaggerate the prevalence of anomalies within a set; so we require 
\[ \E_{\x \sim R|_C}[w(\x)]  \leq \E_{\x \sim R|_C}[w^*(\x)].\]
This captures the soundness requirement in condition (2), if we were to replace the learned weights $w$ by the ground truth $w^*$, the average weights would only increase. 

Equation \eqref{eq:sandwich0} implies the following outer inequality for $w^*$ 
\begin{align}
    \label{eq:outer}
    \E_{\x \sim P|_C}[w^*(\x)] \leq \E_{\x \sim R|_C}[w^*(\x)]. 
\end{align} 
One can show this is a consequence of convexity. This is apparent from the following restatement of Equation \eqref{eq:sandwich0} which is proved in Lemma \ref{lem:sandwich}. Intuitively, this says that we want the weights $w$ to have the right correlation with the true weights $w^*$ under $P|_C$. 
\begin{align}
\label{eq:sandwich2}
    \left(\E_{\x \sim P|_C}[w^*(\x)]\right)^2 \leq \E_{\x \sim P|_C} [w(\x)w^*(\x)] \leq \E_{\x \sim P|_C} \left[ w^*(\x)^2 \right].
\end{align}

Having proposed a notion of set-wise accuracy for importance weights, we now ask how one can find weights that achieve these bounds. We first examine a notion called \macc, which is implicit in the \maxent\ algorithm for learning importance weights \cite{della1997inducing, dudik2004performance, dudik2007maximum}. We will show that this falls short of our goal. We will then propose a stronger notion called \mcab, and show that this indeed gives weights that satisfy the Sandwiching bounds.  

\subsection{\Macc\ and the \maxent\  Algorithm}

Given importance weights $w(x)$ that define a candidate distribution $Q(x) = w(x)P(x)$, a natural property to require is that each set $C\in \mC$ is given similar weight by $Q$ as under $R$. We say that our weights are {\bf multi-accurate} for $\mC$ if $R(C) \approx Q(C)$ for every set $C \in \mC$.
If the weights $w(x)$ are multi-accurate, then $Q$ is indistinguishable from $R$ by $\mC$. To distinguishers that can only perform these tests, the weight function $w$ is just as plausible as the true weight function $w^*$. 

While the term \macc\ is new in this context, this notion is implicit in the MaxEntropy approach to distribution learning, especially the elegant work of \cite{della1997inducing, dudik2004performance, dudik2007maximum}. The set of multi-accurate distributions forms a polytope. The MaxEnt algorithm finds the distribution $Q$ in this polytope which minimizes the KL divergence between $Q$ and $P$, using a (weak agnostic) learning algorithm for $\mC$.  \eat{Equivalently, it finds the distribution $Q(x) =w(x)P(x)$ minimizing $\E_P[w(x)\log(w(x))]$.} Intuitively, minimizing $\KL{Q}{P}$ gives the multi-accurate distribution $Q$ that most closely matches the prior $P$. 

We can reinterpret \macc\ as saying the importance weights are correct on average for every set $C \in \mC$ under $P|_C$ (rather than under $R|_C$). More precisely, \macc\ implies
\begin{align} 
\label{eq:eq-avg}
    \frac{Q(C)}{P(C)} = \E_{\x \sim P|_C}[w(x)] \approx \E_{\x \sim P|_C}[w^*(x)] = \frac{R(C)}{P(C)}
\end{align}
which is why we call it \macc\ in analogy to \cite{hkrr2018,kgz}. This seems similar in form to the lower sandwiching bound. Yet perhaps surprisingly, \macc\ is not strong enough to guarantee sandwiching. Intuitively, while \macc\ guarantees that $w(x)$ and $w^*(x)$ have similar expectations under $P|_C$, it does not guarantee either of the required bounds on the expectation of $w(x)$  {\em under $R|_C$.} This is because  \macc\ does not constrain the distribution of importance weights $w(x)$ within $C$. However Equation \eqref{eq:sandwich2} suggests that sandwiching requires good correlation between $w$ and $w^*$ under $P|_C$. 

Building on this intuition, we construct examples of multi-accurate weights where the desired inequalities in the sandwiching bound can be violated by an arbitrary multiplicative factor. More precisely, for any $B > 1 $ we show instances where \maxent\ returns importance weights $\wme$ such that $\E_{\x \sim P|_C}[w^*(\x)] \geq B\E_{\x \sim R|_C}[\wme(\x)]$ (whereas sandwiching requires $\E_{\x \sim P|_C}[w^*(\x)] \leq \E_{\x \sim R|_C}[w(\x)]$). We show separate examples exhibiting a similar violation of the upper bound (the examples are separate since by Equation \eqref{eq:outer} in every case at least one of the bounds holds). Furthermore, these violations apply to the solution found by \maxent\ (rather than to a contrived distribution that happens to be multi-accurate). \eat{While \maxent\ implies a global upper bound on the weights, we are seeking the sandwiching to hold for all $C$.}

\subsection{Partitions and Multicalibration}

We introduce the notion of multi-calibration for importance weights,  adapting ideas from recent work on multi-group fairness in the supervised setting \citep{hkrr2018} to the unsupervised setting. \Mcab\ is a strengthening of \Macc. We show that multi-calibrated importance weights guarantee sandwiching bounds. 

\eat{

Having identified the limitation of previous approaches to computing importance weights, we define strong generalizations of these properties that we set as objectives. In this work we
Specifically, we will show that by requiring the importance weights that we produce be calibrated for every test set $C \in \mC$, we obtain meaningful lower and upper bounds on the weights within each group. 
 }
 
 {\bf Partitions: } Our first major contribution is to shift from thinking about distributions and importance weights to thinking about partitions of the domain $\X$. A partition $\mS$ of $\X$ is a collection of disjoint subsets $\{S_i\}_{i=1}^m$ whose union is $\X$. To each point $x \in S$, we assign the importance weight $w(x) = w(S) = R(S)/P(S)$. This naturally defines a distribution $Q$ where 
 \[ Q(x) = w(x)P(x) = \frac{R(S)}{P(S)}P(x) = R(S)P(x|S). \]
 This lets us think of $Q$ as a hybrid distribution, where we first sample a state in the partition using $R$, and a point $x \in S$ using $P|_S$. Using the trivial partition of a single set $S_1 =\X$ gives $Q = P$, and at the other extreme using the partition consisting of all singletons gives $Q = R$. 
While it might seem counterintuitive to restrict the family of importance weights while seeking stronger guarantees, it serves to highlight the key challenge in assigning importance weights: finding regions of $\X$ that should receive similar importance weights under $R$. Once we partition space into these regions, the choice of importance weights $R(S)/P(S)$ is natural. 

{\bf \Mcab\ for partitions: }
Now given a set of tests $\mC$, we ask that the importance weights resulting from our partition be calibrated for every $C \in \mC$. While calibration is usually defined for $[0,1]$ valued variables, in our setting it means the following. Consider the set $C \cap S_i$. Every point in this set is assigned the importance weight $w(S_i) = R(S_i)/P(S_i)$. \Mcab\ requires that these weights are indeed correct on average for this set, namely that:
\begin{equation}
\label{eq:intro-avg}
    \E_{\x \sim P|_{S_i \cap C}} [w^*(\x)] = \frac{R(C \cap S_i)}{P(C \cap S_i)} \approx \frac{R(S_i)}{P(S_i)}.
\end{equation}
The power of this definition comes from requiring this condition hold for every $C \in \mC$ and $S_i \in \mS$. 

Before going further, let us see why this is a desirable notion in its own right from a fairness perspective as well as for correctness in a heterogeneous population. Returning to the example of researcher analyzing health data, suppose the weights $w$ arise from a \mcbd\ partition. Let $S_i \in \mS$ be such that $w(S_i) = R(S_i)/P(S_i) = 10$. For any condition  $C$, $C \cap S_i$ is the subset of $C$ that is assigned the weight $10$. By Equation \eqref{eq:intro-avg} \Mcab\ guarantees that average of $w^*$ over these points is close to $10$, thus the weights are justified. Intuitively, this ensures that $w$ is not just correct in expectation over $C$, but that condtioned on $C$, it is well correlated with $w^*$.

Our main technical contribution is showing that the importance weights associated with a \mcab\ partition do indeed satisfy the Sandwiching bounds. This raises the question of efficient computation. We first define a generalization of multi-calibration which can be computed sample efficiently, and which preserves the desirable properties of \mcab. We give an algorithm for computing such a multicalibrated partition, assuming access to a weak agnostic learner for the class $\mC$. The number of states in the output partition is independent of $\mC$. Our algorithm is inspired by the boosting algorithm of Mansour and MacAllester \citep{mansour2002boosting}. Thus under the same learnability assumptions underlying  the MaxEnt algorithm, we get a stronger guarantee.

\subsection{Summary of Our Contributions}
We consider the problem of computing importance weights for a distribution $R$ with respect to a prior $P$, given sample access to both distributions. 
\begin{enumerate}
    \item Based on standard cryptographic assumptions, we argue that point-wise accuracy for importance weights is not possible without making strong assumptions on $P$ and $R$. We propose requiring  set-wise accuracy guarantees for a class $\mC$ of sets that represent statistical tests. We formulate Sandwiching bounds as a notion of set-wise accuracy for importance weights, and show that they capture natural {\em completeness} and {\em soundness} requirements.
    \item We show that the notion of \macc\ inherent in the \maxent\ algorithm does not guarantee Sandwiching bounds, by constructing explicit examples where the bounds are violated.
    \item We introduce the notion of \mcab\ for partitions, inspired by recent work in supervised learning \citep{hkrr2018}. We show that the importance 
    weights resulting from such partitions do guarantee sandwiching bounds.
    \item We define a generalization of multi-calibration that can be computed in a sample-efficient manner. We present an efficient algorithm for constructing such multi-calibrated partitions. 
\end{enumerate}

\subsection*{Outline of this Paper}

In Section \ref{sec:def}, we formally define our notions of \macc\ and \mcab\ for partitions. In Section \ref{sec:sandwich}, we show that \mcab\ gives importance weights that satisfy the sandwiching bounds. We then turn to efficient computation. The original definition of \mcab\ can require extremely high sample complexity. In Section \ref{sec:alpha} we give a relaxed definition of \mcab\ that can be achieved with a limited number of samples. In Section \ref{sec:algorithms}, we provide an efficient algorithm to find a partition that meets the relaxed definition of multi-calibration. We discuss more related work in Section \ref{sec:related}. In Appendix \ref{sec:gap}, we exhibit instances where the multi-accurate distributions found by \maxent\ violate the sandwiching bound, up to arbitrary multiplicative factors. We defer additional proofs to Appendix \ref{sec:proofs} and \ref{sec:sandwich_proof}.

\section{Multi-accurracy and Multi-calibration}\label{sec:def}

We use $[m] =\{1, \ldots, m\}$. We use capitals $(P, Q, R, \cdots)$ to denote distributions and boldface $\x, \y, \cdots$ to denote random variables. We use $\x \sim P$ to denote sampling the variable $\x$ according to distribution $P$, and $\x \sim P|_A$ to denote sampling from $P$ conditioned on an event $A \subseteq \X$, where $P(x|A) = P(x)/P(A)$ for each $x \in A$.  
For two distributions $P, Q$ over $\X$ let 
$\dtv(P,Q) = \sum_{x \in \X}\left|P(x) - Q(x)\right|$.
To every distribution $Q(x)$, one can associate a function $w(x) = Q(x)/P(x)$, which we call the importance weights of $Q$ relative to $P$. 
For any $A \subseteq \X$ observe that
    \begin{equation}
        \label{eq:cond-ex}
        Q(A) = \sum_{x \in A} Q(x) =  \sum_{x \in A} w(x)P(x) = P(A)  \sum_{x \in A} w(x)P(x|A) = P(A) \E_{\x \sim P|_A} [w(x)].
    \end{equation}

In our setting, we have a {\em target} distribution $R$ and a {\em prior} distribution $P$, and denote $w^*(x) = R(x)/P(x)$.  
Our aim is to find importance weights $w(x)$ such that $Q(x) = w(x)P(x)$ is close to $R$. We will consider a  family of sets $\mC$ which could be thought of for instance as decision trees or neural nets of a given depth. 
The indicator functions of sets $C \in \mC$ can be viewed as statistical tests. We will assume there is an efficient algorithm to compute these functions.

\subsection{\Macc\  and \maxent}

In this section, we formally define the notion of multiaccuracy, and prove that it does not guarantee the Sandwiching bounds.

\begin{definition}
\label{def:ae}
    Let $\alpha >0$, let $\mC \subseteq 2^\X$ be a collection of sets. 
    An importance weight function $w:\X \rgta \R$ is $\alpha$-multi-accurate in expectation ($\alpha$-multiAE) for $(P, R, \mC)$ if for every $C \in \mC$, the distribution $Q(x) = w(x)P(x)$ satisfies
    \begin{equation}
        \label{eq:def-ae}
        |Q(C)  - R(C)| \leq \alpha.
    \end{equation}  
\end{definition}

As such, the definition of \macc\ only requires indistinguishability from $R$ for tests in $\mC$.  We might also want $Q$ to be close to $P$ under some divergence, this is  equivalent to minimizing a regularizer term in the weights  \citep{dudik2007maximum}. The \maxent\ algorithm of Dudik, Phillips and Schapire \citep{dudik2004performance, dudik2007maximum}  minimizes $\KL{Q}{P}$ which amounts to using $w\log(w)$ as the regularizer. In the case where $P$ is uniform, this minimization is equivalent to maximizing the entropy of $Q$, hence the name \maxent. Let $\wme$  denote the $\alpha$-multi-accurate importance weight function so that the corresponding distribution $\qme$ minimizes $\KL{Q}{P}$. Then $\wme$ is the optimal solution to a convex optimization problem that has  an efficient algorithm  \citep{dudik2004performance}. The following theorem, proved in Appendix \ref{sec:gap}, shows that the MaxEnt algorithm does not guarantee Sandwiching bounds. 

\begin{theorem}
\label{thm:gap_combined}
For any constant $B > 1$, there exist distributions $P, R$ on $\zo^n$, a collections of sets $\mC$ and $C \in \mC$ such that the \maxent\ algorithm run on $(P, R, \mC)$ with $\alpha =0$ returns a distribution $\qme$ with importance weights $\wme$ such that
\[ \E_{\x \sim P|_C}[w^*(\x)] > B\E_{\x \sim R|_C}[\wme(\x)].\]
Similarly, for any constant $B > 1$, there also exist distributions $P, R$ on $\zo^n$, such that the \maxent\ algorithm run on $(P, R, \mC)$ with $\alpha =0$ finds importance weights $\wme$ such that
\[ \E_{\x \sim R|_C}[\wme(\x)] > B\E_{\x \sim R|_C}[w^*(\x)].\]
\end{theorem}

\eat{
\VS{Do we need these details here?} 
\begin{gather}
    \min \E_{\x \sim P}[w(\x)\log(w(\x))]\\
    \text{s.t. }\E_{\x \sim P}[w(\x)] = 1,\notag\\
    \abs{P(C)\E_{\x \sim P|_C}[w(\x)] - R(C)} \leq \alpha \notag
\end{gather}
It is known that $\wme$ is of the form $\wme(x) = \exp(\sum_i \lambda_i c_i(x))$ where $c_i :\X \rgta \zo$ is the indicator function of $C \in \mC$, hence $\qme$ is a Gibbs distribution over $P$. 
\UW{Perhaps put here the theorem statement that multi-accuracy does not imply Sandwiching}\VS{I agree, given that that section will probably be in the Appendix.}
}

\subsection{Partitions and $\alpha$-\mcab}

A collection of disjoint subsets $\mS = \{S_i\}_{i = 1}^m$ such that $\cup_i S_i = \X$ is called a partition of $\X$ of size $m$. For each $x \in \X$, there exists a unique $S \in \mS$ containing it.  The family of distributions $Q$ we consider are obtained by fixing a partition $\mS$ of $\X$ and then reweighing each $S \in \mS$ so that its weight matches $R$. Within $S$, we retain the marginal distribution $P|_S$. We define this formally below:

\begin{definition}
    Given a prior distribution $P$ and a target distribution $R$ over $\X$, and a partition $\mS$ of $\X$, the $(P, R, \mS)$-reweighted distribution $Q$ over $\X$ is given by
    \begin{equation}
        \label{eq:reweight1}
        Q(x) = R(S)P(x|S) \ \text{for} \ S \in \mS \ s.t. \ x \in S.
    \end{equation}
    Equivalently, the importance weights of $Q$ relative to $P$ are given by
    \begin{equation}
        \label{eq:reweight2}
        w(x) =  \frac{R(S)}{P(S)} \ \text{for} \ S \in \mS \ s.t. \ x \in S.
    \end{equation}
\end{definition}

Since $w$ is constant within each $S \in \mS$, we can view it as a weight function $w: \mS \rgta \R$.  
When $P, R$ and $\mS$ are clear from context, we simply refer to $Q$ as the reweighted distribution. We have the following expressions for any $A \subseteq \X$ and $S \in \mS$,
\begin{align}
\label{eq:q-of-as}
    Q(A \cap S) &= \sum_{x \in A \cap S}R(S) P(x|S) = R(S)P(A|S),\\ 
\label{eq:q-of-a}
    Q(A) &= \sum_{S \in \mS} Q(A \cap S) = \sum_{S \in \mS}R(S)P(A|S).
\end{align}

One can sample from $Q$, given random samples from $R$ and $P$. We first select $S \in \mS$ by drawing a single sample from $R$. We then sample $\x_i \sim P$ until $\x_i \in S$, and return that as our sample from $Q$. Observe that one can interpolate between $P$ and $R$ by making the partition finer. If the partition only consists of the single set $\X$, then $Q = P$. If $\X$ is discrete and we take $\mS$ to be the collection of singleton sets, then $Q = R$. 

Our goal in choosing the partition $\mS$ will be to ensure that the reweighted distribution should be accurate for a set of statistical tests $\mC$, while keeping the number of states small. Note that the class $\mC$ of tests could be large, possibly inifinite (say all halfspaces or neural nets). Our hope is that reweighting a small-sized partition $\mS$ will be sufficient to get accuracy for a large family of tests. We formalize this with the notion of $\alpha$-\mcab.  

\begin{definition}
    \label{def:a-multic}
    ($\alpha$-\mcab) 
    Let $\alpha >0$, let $\mC \subseteq 2^\X$ be a collection of sets. A partition $\mS$ of $\X$ is $\alpha$-multi-calibrated for $(P, R, \mC)$ if for every $C \in \mC$ and $S \in \mS$, the $(P, R, \mS)$-reweighted distribution $Q$ satisfies
    \begin{align}
        \label{eq:big1} \abs{Q(C \cap S)  - R(C \cap S)} & \leq \alpha R(S).
    \end{align}
    Equivalently, $\mS$ is $\alpha$-multi-calibrated for $(P, R, \mC)$ if for every $C \in \mC$ and $S \in \mS$, it holds that
    \begin{align}
        \label{eq:big2} \abs{P(C|S)  - R(C|S)} & \leq \alpha.
    \end{align}
\end{definition}

Let us show that the two formulations are indeed equivalent. By Equation \eqref{eq:q-of-as} we have
$Q(C \cap S) = R(S)P(C|S)$.  Since $R(C \cap S) = R(S)R(C|S)$, we substitute these in Equation \eqref{eq:big1} and divide by $R(S)$ to derive Equation \eqref{eq:big2}. Some observations about the definition:

\begin{itemize}
\item {\bf $\alpha$-multi-calibration implies $\alpha$-multi-accuracy}. 
Multi-calibration for $\mS$ implies multi-accuracy in expectation for the distribution $Q$; this follows easily by summing Equation \eqref{eq:big1} over all $S \in \mS$. Note that we have not used an explicit regularizer to ensure that $Q$ is close to $P$. This condition is instead enforced by keeping the number of states $m$ small, and only allowing the natural weights $w(S) = R(S)/P(S)$.  

\item {\bf Multi-calibration is symmetric in the distributions.} 
If the partition $\mS$ is $\alpha$-multi-calibrated for $(P, R, \mC)$, it is also  $\alpha$-multi-calibrated for $(R, P, \mC)$. This is clear from Equation \eqref{eq:big2} which is symmetric in the two distributions. But note that the $(P, R, \mS)$ reweighted distribution $Q$ and the $(R, P, \mS)$- reweighted  distribution $Q'$ that Equation \eqref{eq:big1} refers to are different.
\end{itemize}

The following technical lemma will be used in proving Sandwiching. 
We think of the importance weights $w$ of $Q$ and $w^*$ of $R$ relative to $P$ as random variables under the distribution $P$ and compare their conditional expectations for each $C \in \mC$.
Readers familiar with the definitions of multi-accuracy and multi-calibration to the corresponding notions in the supervised setting from \cite{hkrr2018} will notice the similarity. The proof is in Appendix \ref{sec:proofs}. 

\begin{lemma}
\label{lem:basic_properties}
    The weight function $w:\X \rgta \R$ is $\alpha$-multiAE for $(P, R, \mC)$ iff for every $C \in \mC$,
    \begin{equation}
        \label{eq:ae}
        \abs{ \E_{\x \sim P|_C}[w(x)] -  \E_{\x \sim P|_C}[w^*(x)] }\leq \frac{\alpha}{P(C)}.
    \end{equation} 
    The partition $\mS$ is $\alpha$-multi-calibrated for $(P, R, \mC)$ iff for every $C \in \mC$ and $S \in \mS$,
    \begin{equation}
        \label{eq:multic}
        \left\lvert  w(S) -  \E_{\x \sim P|_{C \cap S}}[ w^*(x)] \right\rvert \leq \frac{\alpha R(S)}{P(C \cap S)}.
    \end{equation}
    
\end{lemma}

\section{\Mcab\ implies Sandwiching Bounds}
\label{sec:sandwich}

In this section, we assume that weight function $w$ comes from a partition $\mS$ that is $\alpha$-\mcab\ for $(P, R, \mS)$. 
For the  weight function $w$, and $k \geq 1$ define the quantity 
\[ \norm{w}_k = \lb \E_{\x \sim P}[w(\x)^k]\rb^{1/k} = \lb \sum_{S \in \mS}\frac{R(S)^k}{P(S)^{k-1}}\rb^{1/k}. \] 
Observe $\|w\|_1 = 1$ and $\|w\|_k$ increases with $k$. 
The main result of this Section is the following:
\begin{theorem}
\label{thm:sandwich}
    If the partition $\mS$ is $\alpha$-\mcbd\ for $(P, R, \mC)$ and  $w:\X \rgta \R$ is the corresponding importance weight function, then
    \begin{align}
    \label{eq:main-sandwich}
        \E_{\x \sim P|_C}[w^*(\x)] - 2\alpha\frac{\norm{w}_2^2}{R(C)} \leq \E_{\x \sim R|_c}[w(\x)] \leq \E_{\x \sim R|_C}[w^*(\x)] + 3\alpha\frac{\norm{w}_2^2}{R(C)} 
    \end{align}
\end{theorem}

As a first step, we prove the equivalence of the two formulations of the sandwiching bounds that were presented in the introduction. 

\begin{lemma}
\label{lem:sandwich}
Equations \eqref{eq:sandwich0} and \eqref{eq:sandwich2} are equivalent.
\end{lemma}
\begin{proof}
One can rewrite the expectations under $R|_C$ in Equation \eqref{eq:sandwich0} in terms of expectations under $P|_C$ as follows 
\begin{align}
    \E_{\x \sim R|_C}[ w(\x)] &= \frac{\sum_{x \in C}  p(x)w(x)w^*(x)}{\sum_{x \in C} p(x)w^*(x)} = \frac{\E_{\x \sim P|_C}[w(\x)w^*(\x)]}{\E_{\x \sim P|_C} [w^*(x)]}\label{eq:magic1}\\
    \E_{\x \sim R|_C}[w^*(\x)] &= \frac{\sum_{x \in C}  p(x)w^*(x)^2}{\sum_{x \in C} p(x)w^*(x)} = \frac{\E_{\x \sim P|_C}[w^*(\x)^2]}{\E_{\x \sim P|_C} [w^*(x)]}\label{eq:magic2}.
\end{align}
Plugging these into Equation \eqref{eq:sandwich0}, we can rewrite those inequalities as
\begin{align*}
    \E_{\x \sim P|_C}[w^*(\x)] \leq \frac{\E_{\x \sim P|_C}[w(\x)w^*(\x)]}{\E_{\x \sim P|_C} [w^*(x)]} \leq \frac{\E_{\x \sim P|_C}[w^*(\x)^2]}{\E_{\x \sim P|_C} [w^*(x)]}
\end{align*}
which is equivalent to Equation \eqref{eq:sandwich2}. 
\end{proof}

We now proceed with the proof. Using the formulation in Equation \eqref{eq:sandwich2}, we will analyze $\E_{\x \sim P|_C}[w(\x)w^*(\x)]$. The key steps are the next two technical lemmas whose proofs are in the appendix.

\begin{lemma}
\label{lem:tech1}
We have
\begin{align}
\label{eq:tech1} 
    \abs{\E_{\x \sim P|_C}[w(\x)w^*(\x)] - \E_{\s \sim P|_C}[w(\s)^2]} \leq \frac{\alpha \norm{w}_2^2}{P(C)}.
\end{align}
\end{lemma}

\begin{lemma}
\label{lem:tech2}
    We have 
    \begin{align}
    \label{eq:tech2}
            \left(\E_{x\sim P|_C}[w^*(\x)]\right)^2 - 2\alpha \frac{R(C)}{P(C)} \leq \E_{\s \sim P|_C} [w(\s)^2] \leq \E_{\x \sim P|_C}[w^*(\x)^2] + 2\alpha\frac{\norm{w}_2^2}{P(C)}
    \end{align}
\end{lemma}

We now put these together to prove Theorem \ref{thm:sandwich}.

\begin{proof}[Proof of Theorem \ref{thm:sandwich}]
We claim the following inequalities hold
    \begin{align}
    \label{eq:other-sandwich}
         \left(\E_{\x \sim P|_C}[w^*(x)]\right)^2 - \alpha \frac{\norm{w}_2^2 + R(C)}{P(C)} \leq \E_{\x \sim P|_C}[w(\x)w^*(\x)] \leq \E_{\x \sim P|_C} [w^*(x)^2] + 3\alpha \frac{\norm{w}_2^2}{P(C)}.  
    \end{align}
These are an immediate consequence of Lemma \ref{lem:tech1} showing that  $\E_{\x \sim P|_C}[w(\x)w^*(\x)]$ and $\E_{\s \sim P|_C}[w(\s)^2]$ are close, and Lemma \ref{lem:tech2} which gives a sandwiching bound for $\E_{\s \sim P|_C}[w(\s)^2]$.  

Equation \eqref{eq:other-sandwich}  equivalent to Equation \eqref{eq:main-sandwich}. To see this, we use the following equalities from Equation \eqref{eq:magic1} and \eqref{eq:magic2}:
\begin{align*}
    \E_{\x \sim P|_C}[w(\x)w^*(\x)] &= \E_{\x \sim R|_C}[ w(\x)]\E_{\x \sim P|_C} [w^*(\x)], \\
    \E_{\x \sim P|_C}[w^*(\x)^2] &= \E_{\x \sim R|_C}[w^*(\x)]\E_{\x \sim P|_C} [w^*(\x)].
\end{align*}
We plug these into Equation \eqref{eq:other-sandwich} and divide throughout by
$\E_{\x \sim P|_C}[w^*(\x)] = R(C)/P(C)$
to derive Equation \eqref{eq:main-sandwich} and complete the proof.
\end{proof}
\section{$(\alpha, \beta)$-\mcab}
\label{sec:alpha}

Multi-calibration requires the closeness of $R(C|S)$ and $P(C|S)$. However it might be the case that one of $P$ and $R$ assign very little probability to some set $S$.  Since in our model, we only get random samples from $P$ and $R$, enforcing the condition required for multi-calibration might be very expensive in terms of sample complexity, which might depend polynomially on $\max_x(R(x)/P(x), P(x)/R(x))$. This motivates a relaxation that we call $(\alpha, \beta)$-multi-calibration. 

\begin{definition}
\label{def:ab-multic}    
Let $\alpha \geq 0, \beta \geq 0$, let $\mC \subseteq 2^\X$ be a collection of sets. The partition $\mS = \{S_1, \ldots, S_m, T_0, T_1\}$ is $(\alpha, \beta)$-multi-calibrated for $\mC$ under $P, R$ if (1) $P(T_0) \leq \min(\beta, R(T_0))$, (2) $R(T_1) \leq \min(\beta, P(T_1))$, (3) for every $C \in \mC$ and $i \in [m]$,
    \begin{align}
        \label{eq:big} 
        \abs{Q(C \cap S_i)  - R(C \cap S_i)} & \leq \alpha R(S_i).
    \end{align}

\end{definition}

$(\alpha, \beta)$-multi-calibration permits two {\em exceptional} subsets $T_0$ and $T_1$ that do not satisfy Equation \eqref{eq:big1}, but these subsets must have small measure under $P$ and $R$ respectively. We will use $\mT = \{T_0, T_1\}$ to denote the exceptional sets. 
Intuitively, we think of $T_0$ as a region that has small measure under $P$, although $R(T_0)$ could be large. It is hard to ensure that Equation \eqref{eq:big} is met for $T_0$  since samples from $P$ seldom lie in it. Similarly, we allow a region $T_1$ where $R$ allocates low probability, but $P(T_1)$ could be large. 
The advantage of allowing for $\mT$ is that for every $S \in \mS\setminus \mT$, we {\em can assume} that $w(S) = R(S)/P(S)$ is bounded by $O(1/\beta)$. This intuition is formalized in the following lemma. 

\begin{lemma}
\label{lem:bounded}
    Let $T \subseteq \X$ and $c \geq 1$ be such that $R(T)/P(T) \geq c/\beta$. Then $P(T) \leq \beta/c$.
\end{lemma}
\begin{proof}
    Since $R(T)/P(T) \geq c/\beta$, we have $P(T) \leq \beta R(T)/c \leq \beta/c$. 
\end{proof}

When $\beta = 0$ we recover the notion of $\alpha$-multi-calibration. Analogously when $\beta > 0$, we show that the distributions $R$ and $P$ are $\beta$-close in statistical distance to distributions $R^h$ and $P^h$  respectively that are indeed $\alpha$-multicalibrated. 

\begin{definition}
\label{def:hybrid}
    Define the distribution $P^h$ which is identical to $P$ on $\mS\setminus \{T_0 \}$. Let $P^h(T_0) = P(T_0)$, and $P^h|_{T_0} = R|_{T_0}$. Similarly, define $R^h$ to be identical to $R$ on $\mS\setminus\{T_1\}$. Let $R^h(T_1) = R(T_1)$, and $R^h|_{T_1} = P|_{T_1}$.  
\end{definition}

\begin{lemma}
\label{lem:impure}
If the partition $\mS$ is $(\alpha, \beta)$-multi-calibrated for $(P, R, \mC)$, 
then 
\begin{itemize}
    \item $\dtv(P^h,P) \leq \beta$, $\dtv(R^h, R) \leq \beta$.
    \item The partition $\mS$ is $\alpha$-multi-calibrated for $(P^h, R^h, \mC)$. 
\end{itemize}
\end{lemma}
\begin{proof}
    The statistical distance bounds hold since $P^h$ and $P$ only differ on $T_0$ and $P^h(T_0) = P(T_0) \leq \beta$. We verify that the partition is multi-calibrated by showing that $|P(C|S) - R(C|S)| \leq \alpha$ for every state $S \in \mS$.  For any $i \in [m]$, we have 
    \[ \abs{R^h(C|S_i) - P^h(C|S_i)} =  \abs{R(C|S_i) - P(C|S_i)} \leq \alpha. \]
    where the equality holds since since  $P^h$ and $P$ (and $R^h$ and $R$) are identical on the states $S_i$ for $i \in [m]$ and the inequality is from Equation \eqref{eq:big}. The conditional distributions $P^h|_{T_0}$ and $R^h|_{T_0}$ are identical since they both equal $R|_{T_0}$ by construction. Hence $R^h(C|T_0) = P^h(C|T_0)$ for all $C \in \mC$, so the condition holds.  A similar argument holds for $T_1$. 
\end{proof}

A corollary is that $(\alpha, \beta)$-\mcab\ implies $(\alpha + 2\beta)$-\macc. 
\begin{lemma}
\label{lem:mc-ae}
    If $\mS$ is $(\alpha, \beta)$-\mcbd\ for $(P, R, \mC)$, then  the $(P, R, \mS)$-reweighted distribution $Q$ is $\gamma = (\alpha + 2\beta)$-\mact\ for $(P, R, \mC)$.
\end{lemma}

Finally, we can show a sandwiching bound for $(\alpha, \beta)$-\mcab. The error terms now also depend on $\beta$ and $\norm{w}_4^2$ in comparison to Theorem \ref{thm:sandwich}.

\begin{theorem}
\label{thm:sandwich2}
    Assume the partition $\mS$ is $(\alpha, \beta)$-\mcbd\ for $(P, R, \mC)$ and  $w:\X \rgta \R$ is the corresponding importance weight function. Let
    \begin{align} 
    \ell(\alpha, \beta, w) = \alpha\norm{w}_2^2 + \sqrt{\beta}\norm{w}_4^2.
    \end{align}
    Then for every $C \in \mC$, 
    \begin{align}
    \label{eq:main-sandwich2}
        \E_{\x \sim P|_C}[w^*(\x)] - \frac{2\ell(\alpha, \beta, w)}{R(C)} - \frac{2(\alpha +2 \beta)}{P(C)} \leq \E_{\x \sim R|_c}[w(\x)] \leq         \E_{\x \sim R|_C}[w^*(\x)] + \frac{3\ell(\alpha, \beta, w)}{R(C)}.
    \end{align}
\end{theorem}

The proof appears in Appendix \ref{sec:sandwich_proof}.

\eat{

\subsection*{Terminology}
Henceforth \mcab\ will refer to $(\alpha, \beta)$-\mcab. We will refer to the $\beta =0$ case of $\alpha$-\mcab\ as {\em pure} multi-calibration, in analogy with the terminology used in differential privacy.  We summarize the setup and notation we work with below.

\begin{enumerate}
    \item  Let $\alpha, \beta >0$, and let $\gamma = \alpha + 2\beta$. 
    
    \item Let $\mC \subseteq 2^\X$ and $\mS = \{S_1, \ldots, S_m, T_0, T_1\}$ be a partition of $\X$. Let $P$ and $R$ be distributions on $\X$. The partition  $\mS$ is $(\alpha, \beta)$-multi-calibrated for $(P, R, \mC)$ and $Q$ is the $(P, R, \mS)$-reweighted distribution. We use $\mT = \{T_0, T_1\}$ to denote the exceptional sets.
    
    \item We use $P^h$  and $R^h$ to denote the hybrid distributions as in Definition \ref{def:hybrid}.
    
    \item The importance weights of $R$ and $Q$ realtive to $P$ are denoted by $w^*$ and $w$ respectively.  
\end{enumerate}
}
\section{Algorithm for $(\alpha,\beta)$-\mcab\ }\label{sec:algorithms}

In this section, we give an efficient algorithm  that computes a multicalibrated partition, given access to a weak agnostic learner for $\mC$. The algorithm is reminiscent of the algorithm for Boosting via Branching Programs due to \cite{mansour2002boosting}.  We first define weak agnostic learning. Given a collection of sets $\mC \subseteq 2^\X$, we can associate every set $C$ with its indicator function $c:\X \rgta \zo$. 

\begin{definition}
\label{def:weak-ag}
 A $(\alpha, \alpha', L)$-weak agnostic learning algorithm for a class $\mC$ is given $L$ samples from a distribution $\D = (\x, \y)$ where $\x \in \X$ and $\y \in \zo$. If there exists $c \in \mC$ such that $\Pr_\D{[c(\x) = \y}] \geq (1 + \alpha)/2$, then the learner will return $c' \in \mC'$ such that $\Pr_\D[c'(\x) = \y] \geq (1 + \alpha')/2$ for some $0 < \alpha' \leq \alpha$. The class $\mC$ is said to be $(\alpha, \alpha', L)$-weakly agnostically learnable if such a leaner exists. 
\end{definition}
We allow $\alpha - \alpha'$ to depend on $L$, typically it decreases with $L$. For simplicity, we will assume that $\mC' = \mC$, and do not allow for probability of error. Given two distributions $P$ and $R$, a weak learner for $\mC$ can be used to find $C \in \mC$ such that $|R(C) - P(C)|$ is large, a view that we will use hereafter.

\begin{lemma}
    \label{lem:dist}
    Let $\mA$ be an $(\alpha, \alpha', L)$-weak agnostic learner for $\mC$.
    Given distributions $P$ and $R$, if there exists $C \in \mC$ so that $|R(C) - P(C)| \geq \alpha$, given $O(L)$ samples from each of $R$ and $P$, $\mA$ can be used to find $C' \in \mC$ such that $|R(C') - P(C')| \geq \alpha'$.
\end{lemma}
\begin{proof}
    Assume that $R(C) > P(C)$. 
    Define a distribution $\D$ where we output $(\x \sim P, 0)$ or $(\x \sim R, 1)$ each with probability $1/2$. For any $c \in \mC$ we have
    \begin{align*}
        \Pr_{(\x, \y) \sim \D}[c(\x) = y] = \frac{1}{2}\left(\Pr_{\x \sim P}[c(x) =0]   +  \Pr_{\x \sim R}[c(\x) =1]\right) = \frac{1 - P(C) + R(C)}{2} = \frac{1}{2} + \frac{R(C) - P(C)}{2}.
    \end{align*}
    If there exists $C \in \mC$ such that $R(C) - P(C) \geq \alpha$, $\mA$  run on $\D$ will return $C' \in \mC$ such that $R(C') - P(C') \geq \alpha'$. To generate $L$ samples from $\D$, it suffices to have $O(L)$ samples from both  $R$ and $P$.  
\end{proof}

Our algorithm for achieving $(\alpha,\beta)$-multi-calibration uses the weak learners to find distinguishing sets. We next describe our algorithm.

\subsection{Algorithm for Multi-Calibration}

Given sample access to distributions $P, R$, we will construct a multicalibrated partition by starting from the trivial partition and iteratively modifying it till we achieve multi-calibration. We assume access to a weak agnostic learner for the class $\mC$.  

We use $(\mS^t, \mT_0, \mT_1)$ to denote the $t^{th}$ partition, and $Q^t$ to denote the corresponding reweighted distribution. The partition consists of three groups of sets:
\begin{itemize}
    \item {\bf Large weights: } $\mT_0$ consisting of sets $T$ such that $R(T)/P(T) \geq 2/\beta$. 
    
    \item {\bf Small weights: } $\mT_1$ consisting of sets $T$ such that $R(T)/P(T) \leq \beta/2$.
    
    \item {\bf Medium weights: } $\mS^t$ will consists of sets $S$ such that $R(S)/P(S) \in [\beta/2, 2/\beta]$.  
\end{itemize}

The collections $\mT^0, \mT^1$ both start empty and grow monotonically. Once set $T$ is added to either, that set is not modified. All sets in $\mT_0$ will eventually be merged into a single set $T_0$ such that $P(T_0) \leq \beta$, while the sets in $\mT_1$ will be merged into a single set $\mT_1$. Doing the merging at the end simplifies the analysis, but for intuition it is fine to think of each as a single state that keeps growing.

Our algorithm will mostly focus on the medium sets in $\mS^t$, although occasionally sets will be added to $\mT_0$ or $\mT_1$, hence we use the superscript $t$ to account for how it changes over iterations. The algorithm combines two basic operations. 

\begin{itemize}
\item $\Split$: This operation takes  $S \in \mS_t$ where $R(S), P(S)$ are sufficiently large, and $C \in \mC$ such that $|P(C|S) - R(C|S)| > \alpha'$, and split $S$ into two states, $C \cap S$ and $\bar{C} \cap S$. We find the pair $S, C$ by running the weak agnostic learner to distinguish the distributions $P|_S$ and $R|_S$. The new sets are classified as small, medium or large.

\item $\Merge$: This operation is applied to $\mS^t$ when the number states in it goes beyond a certain bound.  It merges those states in $\mS^t$ with similar importance weights into a single state, and halves the number of states.
\end{itemize}

We first state and analyze each of these operations. We view both operations as updating the current partition. 

\begin{algorithm}[ht]
	\caption{$\Split(S, C)$}
	\label{alg:split}
	    {\bf Input: }
	    \begin{itemize}
	        \item $S \in \mS_t$ s.t. $R(S) \geq \beta/4m$, $P(S) \geq \beta/4m$.
	        \item $C \in \mC$ s.t. $|R(C|S) - P(C|S)| > \alpha'$.
	    \end{itemize}
	    
	    Replace $S$ with the two states $S_0 = S \cap C$ and $S_1 = S \cap \bar{C}$.
\end{algorithm}

Note that the states $S_0$ and $S_1$ might end up in any of $\mT_0, \mT_1$ or $\mS^{t+1}$ depending on the value of $R(S_i)/P(S_i)$. We will analyze the Split operation using $\KL{Q_t}{P}$ as the potential function.

\begin{lemma}
\label{lem:split}
    We have $\KL{Q_{t +1}}{P} - \KL{Q_t}{P} \geq 4 R(S)\alpha'^2$.
\end{lemma}
\begin{proof}
    Since $Q_{t+1}$ differs from $Q_t$ by splitting $S$ into  $S \cap C$ and $S \cap \bar{C}$, we have
    \begin{align}
        \KL{Q_{t+1}}{P} - \KL{Q_t}{P} &= R(S \cap C)\logf{R(S \cap C)}{P(S \cap C)} + R(S \cap \bar{C})\logf{R(S \cap \bar{C})}{P(S \cap \bar{C}C)} - R(S)\logf{R(S)}{P(S)}\notag\\
        &= R(S \cap C)\logf{R(S \cap C)P(S)}{R(S)P(S \cap C)} + R(S \cap \bar{C})\logf{R(S \cap \bar{C})P(S)}{R(S)P(S \cap \bar{C})} \notag\\
        & = R(S)\left(R(C|S)\logf{R(C|S)}{P(C|S)} + R(\bar{C}|S)\logf{R(\bar{C}|S)}{P(\bar{C}|S)}\right). \label{eq:kl-diff}
    \end{align}
    The expression in braces is the KL divergence between two Bernoulli random variables that are $1$ with probability $R(C|S)$ and $P(C|S)$ respectively. Hence we can apply Pinsker's inequality to get
    \begin{align}
        \label{eq:Pinsker}
        R(C|S)\logf{R(C|S)}{P(C|S)} + R(\bar{C}|S)\logf{R(\bar{C}|S)}{P(\bar{C}|S)}
         \geq \abs{R(C|S) - P(C|S)}^2 \geq 4\alpha'^2.
    \end{align}
    Plugging this into Equation \eqref{eq:kl-diff} gives the desired bound.
\end{proof}

We now describe the merge operation. \eat{Recall that $\infnorm{R/P} \leq B$, so $R(S)/P(S) \in [1/B, B]$.} For  parameters $\delta$ to be specified later, we divide the interval $[\beta/2,2/\beta]$ into $O(\log(1/\beta)/\delta)$
geometric scales of $\exp(\delta)$. $\Merge$ ensures that the number of states is at most $m$. 

\begin{algorithm}[ht]
	\caption{$\Merge(\delta)$}
	\label{alg:merge}
	    {\bf Input: } parameter $\delta$.
	    \begin{enumerate}
        \item Let $m = \lceil \frac{1}{\delta}\logf{4}{\beta^2} \rceil$.
	    \item For each $i \in \{1, \ldots, m\}$:\\
	    $~~~~~~$ Form a new state $S_i$ by merging all states $S' \in \mS_j$ such that 
	    \[ \frac{R(S')}{P(S')} \in  \left(\frac{e^{(i-1)\delta}\beta}{2}, \frac{e^{i\delta}\beta}{2}\right]. \] 
	    \item Let $\mS^{t+1} = \{S_i\}_{i=1}^m$, discarding any empty states.
	    \end{enumerate}
\end{algorithm}
 Unlike $\Split$, $\Merge$ can reduce the KL divergence, but we can  bound the loss.

\begin{lemma}
\label{lem:merge}
    We have $\KL{Q_{t}}{P} - \KL{Q_{t+1}}{P} \leq \delta$.
\end{lemma}
\begin{proof}
    Let $S'_1, \ldots, S'_\ell \in \mS_t$ denote the states that are merged to form $S_i \in \mS_{t+1}$. For each $k \in [\ell]$,
    \begin{align*}
        \frac{R(S'_k)/P(S'_k)}{R(S_i)/P(S_i)}  \leq  e^\delta.
    \end{align*}
    We use this to bound the decrease in potential from $S_i$ as
    \begin{align*}
        \sum_{k=1}^\ell R(S'_k)\logf{R(S'_k)}{P(S'_k)} - R(S_i)\logf{R(S_i)}{P(S_i)} &= \sum_{k=1}^\ell R(S'_k)\left(\logf{R(S'_k)}{P(S'_k)} - \logf{R(S_i)}{P(S_i)}\right)\\
        &= \sum_{k=1}^\ell R(S'_k)\logf{R(S'_k)/P(S'_k)}{R(S_i)/P(S_i)}\\
        & \leq \sum_{k=1}^\ell R(S'_k)\delta = R(S_i)\delta.
    \end{align*}
    The claim follows by summing over all $S_i \in \mS_{t+1}$. 
\end{proof}

We now state and analyze our main algorithm. 

\begin{algorithm}[ht]
	\caption{Multi-Calibrate$(P, R, \mC, \alpha, \beta)$}
	\label{alg:multiC}
	{\bf Inputs:} 
	\begin{itemize} 
	\item Parameters $\alpha, \beta > 0$.
	\item Distributions $P, R$.
	\item A class $\mC$ that is $(\alpha, \alpha', L)$-weakly agnostically learnable.
	\end{itemize}
	{\bf Output: } A partition  that is $(\alpha, \beta)$-multicalibrated for $\mC$ under $P, R$.\\
	
	    Let $\mS^1 = \{\X\}, \mT^1_0 = \mT^1_1 = \{\}$.\\
	    Let $ \delta= \beta\alpha'^2/2$, and $m = \lceil \frac{1}{\delta}\logf{4}{\beta^2} \rceil$.\\
	    For $t \geq 1$
	    \begin{enumerate}
	    \item If $|\mS_t| \geq 2m$, then run $\Merge (\delta)$. 
	    \item If the weak agnostic leaner finds $S \in \mS^t$, $C \in \mC$ such that 
	    \begin{align*} 
	    R(S) \geq \beta/4m, P(S) \geq \beta/4m, \ \abs{R(C|S) - P(C|S)} \geq \alpha'. 
	    \end{align*}
	    \begin{enumerate}
	    \item Run $\Split(S, C)$ and obtain $S_0, S_1$
	    \item If $P(S_0) < \beta/4m$ and $P(S_0) < R(S_0)$, place $S_0$ in $\mT_0$. Else, if $R(S_0) < \beta/4m$ place $S_0$ in $\mT_1$.
	    \item Repeat previous step for $S_1$
	    \item Repeat the loop
	    \end{enumerate}  
	    If the weak learner fails, exit the loop.
	    
        \end{enumerate}
        {\bf Post-Processing: }
	   \begin{enumerate} 
	   \item Move all $S \in \mS^t$ such that $P(S) < \beta/4m$ and $P(S) \leq R(S)$ from $\mS^t$ to $T_0$. Move all remaining $S \in \mS^t$ such that $R(S) < \beta/4m$ from $\mS^t$ to $T_1$. 
	   
	   \item Merge all $T \in \mT_0$ into a single state $T_0$. Merge all $T \in \mT_1$ into a single state $T_1$. 
	   
	   \item Return the partition $\mS = \mS^t \cup \{T_0\} \cup \{T_1\}$.
	   \end{enumerate}
\end{algorithm}

\begin{theorem}
\label{thm:correct}
     Algorithm \ref{alg:multiC} returns a partition $\mS$ that is $(\alpha, \beta)$-\mcbd\ for $\mC$ under $P, R$. 
\end{theorem}
\begin{proof}
    We first prove that $P(T_0) \leq \beta$ and $P(T_0) \leq R(T_0)$. We can write $T_0 = \cup_i T'_i \cup_j S'_j$ where the sets $T'_i$ were added to $T_0$ during the loop, when they were created during a $\Split$ operation, and the sets $S'_j$ were moved from $\mS^{t}$ in the post-processing step. Then $R(T'_j)/P(T'_j) \geq 2/\beta$ for all $j$, hence $R(\cup_j T'_j)/P(\cup_j T'_j) \geq 2/\beta$. But by Lemma \ref{lem:bounded}, this implies that $P(\cup_j T'_j) \leq \beta/2$. The sets $S'_j$ are added to $T_0$ because $P(S'_j) \leq \beta/4m$. Since there are at most $2m$ such sets (else we would have run $\Merge$), we have $P(\cup_j S'_j) \leq 2m \beta/4m \leq \beta/2$.  Overall
    \[ P(T_0) \leq P(\cup_i T'_i) + P(\cup_j S'_j) \leq \beta/2 + \beta/2 = \beta. \] Further, for every set $T$ merged into $T_0$ it holds that $P(T) \leq R(T)$ and therefore $P(T_0) \leq R(T_0)$.
    A similar argument shows that $R(T_1) \leq \beta$ and $R(T_1) < P(T_1)$. 
    
    We need to show that every set $S \in \mS^t$  satisfies $\|R(C|S)  - P(C|S)\| \leq \alpha$ for all $C \in \mC$. Note that  $S$ satisfies $R(S) \geq \beta/4m$ and $P(S) \geq \beta/4m$, else it would have been removed from $\mS^t$ in the post-processing step. Hence, if it violates this condition, the weak agnostic learner would find a $C' \in\mC$ such that $\|R(C'|S)  - P(C'|S)\| \geq \alpha'$, so we would not exit the loop at the $t^{th}$ iteration.  
    
    This shows that the partition $\mS$ is $(\alpha, \beta)$-\mcbd.  
\end{proof}

Next we analyze the running time and sample complexity.
\begin{theorem}
\label{thm:runtime}
     Given an $(\alpha, \alpha', L)$ weak agnostic learning algorithm for $\mC$, Algorithm \ref{alg:multiC} returns an $(\alpha, \beta)$-\mcbd\ partition within  $T$ iterations where 
     $T =  \Tilde{O}(KL(R,P)/(\beta^2 \alpha'^4)$. 
    It makes $O(T)$ calls to the weak agnostic learner, where each call requires $\Tilde{O}(L /(\beta^2 \alpha'^2))$ samples from each of $R$ and $P$.
\end{theorem}
\begin{proof}
    Each iteration but the last involves one call to either $\Split$ or $\Merge$.
    We bound the number of calls to $\Merge$, denoted $\ell$.  Assume the merge operations happen in interations $t^1 < t^2 \cdots < t^\ell)$. 
    Every $\Split$ operation increases the number of states by $1$, whereas $\Merge$ reduces it from $2m$ to a number is the range $\{1, \ldots, m\}$. Hence $2m \leq t^{k+1} -t^{k} \geq m$. Each $\Split$ operation acts on a set $S$ where $R(S) \geq \beta/4m$, and by Lemma \ref{lem:split}, it increase the KL divergence by $4R(S)\alpha'^2$. 
    The $\Merge$ operation decreases it by $\delta = \alpha'^2\beta/2$. Hence we have
    \begin{align*}
        \KL{Q_{t^{k+1}}}{P} - \KL{Q_{t^k}}{P} \geq m \frac{\beta}{4m} 4\alpha'^2 - \delta = \delta.
    \end{align*}
    Thus the KL divergence between successive $\Merge$ operations increases by $\delta$. We start with the trivial partition, so $Q^1 = P$. 
    Since $\mS^T$ is  partition, if $Q^T$ denotes the corresponding reweighted distribution, then $\KL{Q^T}{P} \leq \KL{R}{P}$. Hence
    \[ \ell \delta \leq \KL{Q^T}{P} - \KL{Q^1}{P} \leq  \KL{R}{P}\] 
    hence $\ell \leq \KL{R}{P}/\delta$. The total number of iterations is bounded by 
     \[ T \leq (2m + 1)\ell =  O(\log(1/\beta)\KL{R}{P}/\delta^2) = \Tilde{O}(\KL{R}{P}/(\beta^2\alpha'^4).\]
     
     For one $\Split$ iteration, we might make $O(m)$ calls to the weak learner, one per state to find the pair $S, C$ on which to run $\Split$. However, once  we fail to find a good $C$ for $S$, we do not need to try $S$  again until the state is modified, which cannot happen before the next $\Merge$ iteration. This shows that there are at most $4m$ calls to the agnostic learner between two merge operations, $2m$ successful ones and $2m$ unsuccessful ones. Hence the number of calls to the learner is bounded by $4m\ell = O(T)$.
 
     Finally, we address the sample complexity. 
     We need to run the learner on the distributions $P|_S$ and $R|_S$ where $R(S), P(S) \geq \beta/4m$. If the sample complexity of the agnostic learner is $L$ then $O(Lm/\beta) = \Tilde{O}(L/(\alpha'\beta)^2)$ samples from each of $P$ and $R$ will suffice to ensure that we have sufficiently many samples from $P|_S$ and $R|_S$ respectively. 
    \end{proof}

Finally we note that the partition we compute can be represented by a $
\mC$-branching program where each node is labelled by $c \in \mC$. 

\section{Other Related Work}\label{sec:related}

In Section \ref{sec:intro}, we discussed some of the diverse applications of importance weights which span many communities. We refer the reader to the book \cite{sugiyama2012density} for a more comprehensive overview of the related work, especially in the context of the machine learning literature. Here, we provide a brief overview of some of the important techniques for computing importance weights.

Owing to extensive applications, there has  been considerable interest in showing theoretical guarantees for various approaches for importance weight estimation. An influential work in this regard is \cite{nguyen2010estimating} which shows a variational characterization of KL divergence as an optimization problem over some hypothesis class $\mathcal{H}$, the solution to which also yields the importance weights. The optimization problem is convex if $\mathcal{H}$ is convex, and they establish guarantees on the rate of convergence of the estimated importance weights to the true weights. However, their analysis crucially assumes that the true importance weights lie in the hypothesis class $\mathcal{H}$---which is a strong assumption in most applications. There are also other works which show guarantees on the recovered importance weights if the true importance weights belong to simple hypothesis classes such as linear functions \citep{kanamori2010theoretical,kanamori2009efficient}. Under practical scenarios where the true importance weights are too complex to be modelled by any simple hypothesis class $\mathcal{H}$, these results do not provide any guarantees for the predicted importance weights to be accurate or calibrated. 

To allow more expressive hypothesis classes, kernel based approaches for estimating importance weights have also been proposed, starting with Kernel Mean Matching (KMM) introduced in \cite{huang2007correcting}. If the underlying kernel is universal, then under the limit of infinite data KMM provably recovers the true importance weights \citep{cortes2008sample,huang2007correcting}. In the limit of finite data however, kernel based approaches can be interpreted as generalizations of moment-matching methods such as \citep{qin1998inferences} which seek to match some moments of the data \citep{sugiyama2012adensity}. In the context of our work, this implies that the kernel based approaches guarantee indistinguishability with respect to the matched moments (or more generally, the feature space induced by the kernel), which is equivalent to multi-accuracy in expectation with respect to the moments (or the feature space more generally). 
Also, as we discussed before, a widely studied  body of work which can also be interpreted as getting multi-accuracy guarantees is the concept of {maximum entropy} for distribution estimation \citep{jaynes1957information,phillips2004maximum,dudik2004performance,dudik2007maximum}

\bibliographystyle{alpha}
\bibliography{ref}

\appendix
\section{Gap Instances for \maxent}
\label{sec:gap}

In this section we will show that the importance weights $\wme$ found by \maxent\ need not satisfy the sandwiching bounds. Indeed, for either direction of the sandwiching bound, we will show instances where the inequality is off by an arbitrarily large constant factor. Thus while one would like $\E_{P|_C}[w^*(x)] \leq \E_{R|_C}[w(x)]$, we will exhibit $P, R$ and $\mC$ such that the importance weights $\wme$ found by \maxent\ are such that the ratio
$\E_{P|_C}[w^*(x)]/\E_{R|_C}[w(x)]$ is arbitrarily large, and similarly for the upper bound. Both our counterexamples work by starting with a small example on $\zo^2$ that shows some small constant gap and then tensoring to amplify the gap.

\begin{lemma}
\label{lem:gap-1}
There exist distribution $P, R$ on $\zo^2$, a collections of sets $\mC$ and $C \in \mC$ such that the \maxent\ algorithm run on $(P, R, \mC)$ with $\alpha =0$ returns a distribution $\qme$ with importance weights $\wme$ such that
\[ \E_{\x \sim P|_C}[w^*(\x)] > \E_{\x \sim R|_C}[\wme(\x)].\]
\end{lemma}
\begin{proof}
Let $P$ be the uniform distribution on $\zo^2$. Let $R$ be the distribution where
\[ R(00) = 0, R(01) = R(10) = 3/8, R(11) =1/4.\]
 We denote the two coordinates $x_0, x_1$, and let $\mC$ consist of all subcubes of dimension $1$. Hence $\mC = \{x: x_i =a\}_{i \in \zo, a \in \zo}$. 
 
 The distribution $\qme$ for $\alpha = 0$ is the product distribution which matches the marginal distributions on each coordinate: $\qme(x_0 =1) = \qme(x_1 =1) = 5/8$ and the coordinates are independent. The \macc\ constraints $\qme(x_0 =1) = R(x_0 =1)$ and $\qme(x_1 = 1) = R(x_1 = 1)$ are clearly satisfied, and $\qme$ is the maximum entropy distribution satisfying these constraints. 
 
 We can compute the following importance weights
 \begin{enumerate}
     \item $\wme(11) = (5/8)^2/(1/4)^2 = 25/16$ whereas $w^*(11) = 1$. 
     \item $\wme(10) = (5/8\cdot 3/8)/(1/4)^2 = 15/16$, whereas $w^*(10) = (3/8)/(1/4) = 3/2$; ditto for $01$.    
 \end{enumerate}
 For intuition as to why this is a gap example, note that this shows that while $w^*$ assigns high weights to $01$ and $10$, $\wme$ assigns these points weights less than $1$, and instead assigns a high weight to $11$. Thus an algorithm that was labelling points with $\wme$ exceeding $1$ as anomalies would report $11$ as the sole anomaly, and miss both $01$ and $10$. 
 
 We consider the set $C = \{10, 11\} = \{x: x_0 = 1\}$. Note that $P|_C$ is uniform on $x_1 \in \zo$, whereas $R|_C(x_1 = 1) = 2/5$.  Then it follows that
 \begin{align*}
    \E_{\x \sim R|_C}[\wme(\x)] &= 3/5\cdot 15/16 + 2/5\cdot 25/16 = 19/16\\
    \E_{\x \sim P|_C}[w^*(\x)] &= 1/2\cdot 1 + 1/2\cdot 3/2 = 5/4 
 \end{align*}
 hence $\E_{\x \sim P|_C}w^*(\x) > \E_{\x \sim R|_C}\wme(\x)$. 
 \end{proof}
 
 Intuitively, in the example above, while $\qme$ assigns the right weight of $5/8$ to the set $C$, within $C$ the distribution of weight is misaligned with $R$, leading to low expected weight under $R|_C$. We now tensor this example to amplify the gap.
 
\begin{theorem}
\label{thm:gap-1}
For any constant $B > 1$, there exist distributions $P, R$ on $\zo^n$, a collections of sets $\mC$ and $C \in \mC$ such that the \maxent\ algorithm run on $(P, R, \mC)$ with $\alpha =0$ returns a distribution $\qme$ with importance weights $\wme$ such that
\[ \E_{\x \sim P|_C}[w^*(\x)] > B\E_{\x \sim R|_C}[\wme(\x)].\]
\end{theorem}
\begin{proof}
 We now consider the $k$-wise tensor of the instances constructed in Lemma \ref{lem:gap-1}. The domain is $\zo^{2k}$ where the coordinates are denoted $x_0, \ldots, x_{2k-1}$. We consider the pair of distributions $P_k =(P)^k$ which is uniform on $2k$ bits, $R_k = (R)^k$ which the product of $k$ independent copies of $R$ on the pairs $\{x_{2i}x_{2i+1}\}_{i=1}^{k-1}$. Let $\mC_k$ consist of all subcubes of dimension $k$ where we restrict one co-ordinate out of $x_{2i}, x_{2i +1}$ for $i \in \{0,\ldots, k-1\}$. One can verify that \maxent\  returns $\qme_k = (\qme)^k$ which is just the product distribution on $\zo^{2k}$ with $\Pr[x_i =1] = 5/8$ for every coordinate. 
 
 Let $\wme_k$ and $w^*_k(x)$ denote the importance weights of $\qme$ and $R_k$ with respect to $P_k$. A key observations is that importance weights tensor: for any $x \in \zo^{2k}$, 
 \begin{align*}
     w^*_k(x) &= \frac{R_k(x)}{P_k(x)} = \prod_{i=0}^{k-1} \frac{R(x_{2i}x_{2i +1})}{P(x_{2i}x_{2i +1})} = \prod_{i=0}^{k-1}w^*(x_{2i}x_{2i +1})\\
     \wme_k(x) &= \frac{\qme_k(x)}{P_k(x)} = \prod_{i=0}^{k-1} \frac{\qme(x_{2i}x_{2i +1})}{P(x_{2i}x_{2i +1})} = \prod_{i=0}^{k-1}\wme(x_{2i}x_{2i +1}). 
  \end{align*}
 
  We consider the set $C = \{x: x_{2i} = 1, i \in \{0, \ldots, k-1\}\}$. The key property of this set is that the conditional distributions $P_k|_C = (P|_C)^k$ and $R_k|_C = (R|_C)^k$ are also product distributions of the conditional distributions. Hence we have
  \begin{align*} 
  \E_{\x \sim R_k|_C}[\wme_k(\x)] &= \E_{\x \sim R_k|_C} \left[\prod_{i=0}^{k-1}\wme(\x_{2i}\x_{2i +1})\right] = \prod_{i=1}^{k-1}\E_{\x_{2i}\x_{2i+1} \sim R|_C} [\wme(\x_{2i}\x_{2i +1})] = (19/16)^k\\
  \E_{\x \sim P_k|_C}[w^*_k(\x)] &= \E_{\x \sim P_k|_C} \left[\prod_{i=0}^{k-1}w^*(\x_{2i}\x_{2i +1})\right] = \prod_{i=1}^{k-1}\E_{\x_{2i}\x_{2i+1} \sim P|_C} [w^*(\x_{2i}\x_{2i +1})] = (5/4)^k.
  \end{align*}
  Now take $k$ sufficiently large so that $(5/4)^k > B(19/16)^k$.
  \end{proof}
  
We now construct a gap example for the other direction of the sandwiching bounds, where $\E_{R|_C}[w(x)] > \E_{R_C}[w^*(x)]$. Again we start with a small constant gap and amplify it by tensoring. We will only describe the construction for achieving the small constant gap, the tensoring step is identical to Theorem \ref{thm:gap-1}.

\begin{theorem}
\label{thm:gap-2}
For any constant $B > 1$, there exist distributions $P, R$ on $\zo^n$, a collections of sets $\mC$ and $C \in \mC$ such that the \maxent\ algorithm run on $(P, R, \mC)$ with $\alpha =0$ returns a distribution $\qme$ with importance weights $\wme$ such that
\[ \E_{\x \sim R|_C}[\wme(\x)] > B\E_{\x \sim R|_C}[w^*(\x)].\]
\end{theorem}
\begin{proof}
    As before let $P$ be uniform on $\zo^2$. Consider the distribution $R$ given by
    \[ R(00) = 2/16, R(10) = 6/16, R(01) = 3/16, R(11) = 5/16. \]
    As before we let $\mC$ consist of all subcubes of dimension $1$. The distribution $\qme$ is the product distribution on $x_0$ and $x_1$ where $\Pr[x_0 =1] = 1/2$ and $\Pr[X_1 =1] = 11/16$. We will use the set $C = \{01, 11\}$, so that $R|_C(01) = 3/8, R|_C(11) = 5/8$. 
    
    We compute the importance weights  within $C$ as follows:
    \begin{align*}
        w^*(01) = 3/4, w^*(11) = 5/4\\
        \wme(01) = 5/8, \wme(11) = 11/8. 
    \end{align*}
    Hence we have the conditional expectations
    \begin{align*}
        \E_{\x \sim R|_C}[w^*(\x)] = 3/8\cdot 3/4 + 5/8\cdot 5/4 = 34/32.\\
        \E_{\x \sim R|_C}[\wme(\x)] = 3/8\cdot 5/8 + 5/8\cdot 11/8 = 35/32
    \end{align*}
    hence $\E_{\x \sim R|_C}[\wme(\x)] > \E_{\x \sim R|_C}[w^*(\x)]$. We can amplify this gap by tensoring.
\end{proof}

\section{Additional Proofs}
\label{sec:proofs}

\begin{proof}[Proof of Lemma \ref{lem:basic_properties}]

    Applying Equation \eqref{eq:cond-ex} to distributions $Q$ and $R$ for the set $C$, we get
    \[ Q(C) = P(C) \E_{\x \sim P|_C}[w(x)], \ \  R(C) = P(C)\E_{\x \sim P|_C}[w^*(x)].\]
    Hence we get 
    \[ \abs{Q(C) - R(C)} = P(C) \left\lvert  w(S) -  \E_{\x \sim P|_{C \cap S}}[ w^*(x)] \right\rvert \] 
    and hence  dividing both sides of Equation~\eqref{eq:def-ae} by $P(C)$ gives  Equation~\eqref{eq:ae}.
    
    Applying Equation \eqref{eq:cond-ex} to $R$ with $A = C \cap S$, we get 
    \[ R(C \cap S) = P(C \cap S) \E_{\x \sim P|_{C \cap S}}[ w^*(x)]\]
    whereas by Equation \eqref{eq:q-of-as},
    \[ Q(C \cap S) = R(S)P(C|S) = \frac{R(S)P(C \cap S)}{P(S)} = w(S)P(C \cap S).\] 
    Hence the LHS of Equation~\eqref{eq:big1} can be written as
    \[ \abs{R(C \cap S) - Q(C \cap S)} = P(C \cap S)\left\lvert  w(S) -  \E_{\x \sim P|_{C \cap S}}[ w^*(x)] \right\rvert \]
    We derive Equation~\eqref{eq:multic} by diving both sides of Equation \eqref{eq:big1} by $P(C \cap S)$.
\end{proof}

\begin{proof}[Proof of Lemma \ref{lem:tech1}]
We  sample from the distribution $P|_C$ in two steps:
\begin{enumerate}
    \item We first sample $\s \in \mS$ according to the marginal distribution induced by $P|_C$ where $\Pr[{\s = S_i}] = P(C \cap S_i)/P(C)$.
    \item We then sample $\x \in \s$ according to $P|_{C \cap \s}$ so that $\Pr[\x = x] = P(x)/P(C \cap \s)$. 
\end{enumerate}
This allows us to use the fact that $w(x) = w(\s)$ remains constant within each set of the partition, and that \mcab\ implies that $\E_{P|_{\s \cap C}}[w^*(x)]$ is close to $w(\s)$ by Equation \eqref{eq:multic}. 
\begin{align*}
    \E_{\x \sim P|_C}[w(\x)w^*(\x)] &= \E_{\s \sim P|_C}\E_{\x \sim P|_{\s \cap S}} [w(\x)w^*(\x)] =  \E_{\s \sim P|_C}w(\s)\E_{\x \sim P|_{\s \cap C}} [w^*(\x)].
\end{align*}

 Hence using Equation \eqref{eq:multic} we have 
\begin{align*}
    \abs{\E_{\x \sim P|_C}[w(\x)w^*(\x)] - \E_{\s \sim P|_C}[w(\s)^2]}  &= 
    \abs{\E_{\s \sim P|_C}\left[w(\s)\E_{\x \sim P|_{\s \cap C}} [w^*(\x)] - w(\s)^2\right]}\\
    & \leq  \E_{\s \sim P|_C}\left[w(\s)\abs{\E_{\x \sim P|_{\s \cap C}} [w^*(\x)] - w(\s)}\right]\\
    &\leq \E_{\s \sim P|_C}\left[w(\s)\frac{\alpha R(\s)}{P(\s \cap C)}\right]\\
    & = \sum_{S \in \mS} \frac{P(S \cap C)}{P(C)} \frac{R(S)}{P(S)} \frac{\alpha R(S)}{P(S \cap C)}\\
    &=  \sum_{S \in \mS} \frac{\alpha R(S)^2}{P(S)P(C)} =  \frac{\alpha \|w\|^2}{P(C)}.
\end{align*}
\end{proof}

\begin{proof}[Proof of Lemma \ref{lem:tech2}]
We start with the lower bound.
By Equation \eqref{eq:multic} we have
\begin{align*}
    \E_{\s \sim P|_C}[w(\s)] &\geq \E_{\s \sim P|_C} \left[\E_{\x \sim P|_{\s \cap C}}[w^*(\x)] - \frac{\alpha R(S)}{P(C \cap S)}\right] = \E_{\x \sim P|_C} [w^*(\x)]  - \frac{\alpha}{P(C)}.
\end{align*}
Using this bound and the convexity of $x^2$,
\begin{align*}
    \E_{\s \sim P|_C} [w(\s)^2]  \geq \left(\E_{\s \sim P|_C}[w(\s)]\right)^2    \geq \left(\E_{\x \sim P|_C}[w^*(\x)]- \frac{\alpha}{P(C)}\right)^2 \geq \left(\E_{x\sim P|_C}[w^*(\x)]\right)^2 - 2\alpha \frac{R(C)}{P(C)}.
\end{align*}

We now show the upper bound. 
\begin{align}
\label{eq:lb-w}
    \E_{\x \sim P|_C}[w^*(\x)^2] &= \E_{\s \sim P|_C} \E_{\x \sim P|_{\s \cap C}}[w^*(\x)^2] \geq \E_{\s \sim P|_C} \left( \E_{\x \sim P|_{\s \cap C}}[w^*(\x)] \right)^2\notag\\
    & \geq \E_{\s \sim P|_C} \left[\left(w(S) - \frac{\alpha R(S)}{P(S \cap C)}\right)^2\right]\notag\\
    & \geq \E_{\s \sim P|_C}\left[w(S)^2 - 2w(S)\frac{\alpha R(S)}{P(S \cap C)}\right].
\end{align}
We have
\begin{align*}
\E_{\s \sim P|_C}\left[w(S)\frac{\alpha R(S)}{P(S \cap C)}\right] &= \sum_{S \in \mS} \frac{P(S \cap C)}{P(C)} \frac{R(S)}{P(S)}\frac{\alpha R(S)}{P(S \cap C)} = 
\sum_{S \in \mS}\alpha \frac{R(S)^2}{P(S)P(C)} = \frac{\alpha \|w\|^2}{P(C)}. 
\end{align*}
Plugging this into Equation \eqref{eq:lb-w} gives
\begin{align*}
    \E_{\x \sim P|_C}[w^*(\x)^2] \geq \E_{\s \sim P|_C}[w(S)^2] - 2\frac{\alpha \|w\|^2}{P(C)}
\end{align*}
which gives the desired upper bound.
\end{proof}

\begin{proof}[Proof of Lemma \ref{lem:mc-ae}]
    Fix any $C \in \mC$. Since $\mS$ is $\alpha$-multi-calibrated for $(P^h, R^h, \mC)$ and $\alpha$-\mcab\ implies $\alpha$-\macc, $|P^h(C) - R^h(C)| \leq \alpha$.
    Also from Lemma \ref{lem:impure}, 
    \[ \abs{R(C) - R^h(C)} \leq \dtv(R, R^h) \leq \beta, \ \ \abs{P(C) - P^h(C)} \leq \dtv(P, P^h) \leq \beta. \]
    Now using the triangle inequality,
    \begin{align*}
        \abs{R(C) - P(C)} \leq  \abs{R^h(C) - P^h(C)} + \abs{R(C) - R^h(C)} + \abs{P^h(C) - P(C)} \leq \alpha + 2\beta.
    \end{align*}
\end{proof}

\section{Sandwiching for $(\alpha, \beta)$-\mcab}\label{sec:sandwich_proof}

In this section, we prove Theorem \ref{thm:sandwich2} which asserts that for  
\[ \ell(\alpha, \beta, w) = \alpha\norm{w}_2^2 + \sqrt{\beta}\norm{w}_4^2 \]
the following bounds hold for every $C \in \mC$,
\begin{align*}
    \E_{\x \sim P|_C}[w^*(\x)] - \frac{2\ell(\alpha, \beta, w)}{R(C)} - \frac{2(\alpha +2 \beta)}{P(C)} \leq \E_{\x \sim R|_c}[w(\x)] \leq         \E_{\x \sim R|_C}[w^*(\x)] + \frac{3\ell(\alpha, \beta, w)}{R(C)}.
\end{align*}

Throughout this section, we assume that $\mS = \{S_1,\ldots, S_m, T_0, T_1\}$ is $(\alpha, \beta)$-\mcab\ for $(P, R, \mC)$. We will use $S \in \mS$ to denote a generic set in the partition, which could one of the $S_i$s or $T_j$s. We will now prove a sequence of technical lemmas that will be used to prove our bounds.

\begin{lemma}
\label{lem:s-upper}
For all $C \in \mC$, we have
\begin{align}
\label{eq:s-upper}
    \sum_{i \in [m]}\frac{P(S_i \cap C)}{P(C)}\lb \frac{R(S_i \cap C)^2}{P(S_i \cap C)^2} - \frac{R(S_i)R(S_i \cap C)}{P(S_i)P(S_i \cap C)} \rb & \geq -\frac{3\alpha \norm{w}_2^2}{P(C)},\\
\label{eq:s-lower}    
    \sum_{i \in [m]}\frac{P(S_i \cap C)}{P(C)}\lb\frac{R(S_i)R(S_i \cap C)}{P(S_i)P(S_i \cap C)} - \frac{R(S_i)^2}{P(S_i)^2}\rb &\geq - \frac{\alpha \norm{w}_2^2}{P(C)}.
\end{align}
\end{lemma}
\begin{proof}
Using the \mcab\ condition (Equation \eqref{eq:big2}), we have
\begin{align*}
    \frac{R(S_i \cap C)^2}{P(S_i \cap C)^2} &\geq \lb \frac{R(S_i)}{P(S_i)} - \alpha \frac{R(S_i)}{P(S_i \cap C)} \rb^2 \geq \frac{R(S_i)^2}{P(S_i)^2} - 2\alpha \frac{R(S_i)^2}{P(S_i)P(S_i \cap C)}\\
    \frac{R(S_i)R(S_i \cap C)}{P(S_i)P(S_i \cap C)} &\leq  \frac{R(S_i)}{P(S_i)} \lb \frac{R(S_i)}{P(S_i)} + \alpha\frac{R(S_i)}{P(S_i \cap C)} \rb = \frac{R(S_i)^2}{P(S_i)^2} +\alpha \frac{R(S_i)^2}{P(S_i)P(S_i \cap C)}.
\end{align*}
Subtracting the two bounds and averaging over $S_i$s we get
\begin{align*}
    \sum_{i \in [m]}\frac{P(S_i \cap C)}{P(C)}\lb \frac{R(S_i \cap C)^2}{P(S_i \cap C)^2} - \frac{R(S_i)R(S_i \cap C)}{P(S_i)P(S_i \cap C)} \rb &\geq -3\alpha \sum_{i \in [m]}\frac{P(S_i \cap C)}{P(C)} \frac{R(S_i)^2}{P(S_i)P(S_i \cap C)}\\
    &= -\frac{3\alpha}{P(C)} \sum_{i \in [m]}\frac{R(S_i)^2}{P(S_i)}\\
    & \geq -\frac{3\alpha}{P(C)}\norm{w}_2^2
\end{align*}
which proves Equation \eqref{eq:s-upper}.

We now prove \eqref{eq:s-lower}. By the \mcab\ condition,
\begin{align*}
\sum_{i \in [m]}\frac{P(S_i \cap C)}{P(C)}\frac{R(S_i)R(S_i \cap C)}{P(S_i)P(S_i \cap C)} &\geq \sum_{i \in [m]}\frac{P(S_i \cap C)}{P(C)}\frac{R(S_i)}{P(S_i)}\lb \frac{R(S_i)}{P(S_i)} - \alpha \frac{R(S_i)}{P(S_i \cap C)}\rb \\
&=  \sum_{i \in [m]} \frac{P(S_i \cap C)}{P(C)}\frac{R(S_i)^2}{P(S_i)^2} - \sum_{i \in [m]}\frac{R(S_i)^2}{P(C)P(S_i)}.
\end{align*}
Hence 
\begin{align*}
\sum_{i \in [m]}\frac{P(S_i \cap C)}{P(C)}\lb\frac{R(S_i)R(S_i \cap C)}{P(S_i)P(S_i \cap C)} - \frac{R(S_i)^2}{P(S_i)^2}\rb \geq - \frac{\alpha \norm{w}_2^2}{P(C)}.
\end{align*}
\end{proof}

Next we consider the $T_0$ term and show the following bounds
\begin{lemma}
\label{lem:t0-upper}
For all $C \in \mC$, we have
 \begin{align}
    \label{eq:t0-upper}
    \frac{P(T_0 \cap C)}{P(C)} \lb \frac{R(T_0 \cap C)^2}{P(T_0 \cap C)^2} - \frac{R(T_0)R(T_0 \cap C)}{P(T_0)P(T_0 \cap C)}\rb \geq -\frac{\sqrt{\beta}\norm{w}_4^2}{P(C)},\\
    \label{eq:t0-lower}
    \frac{P(T_0 \cap C)}{P(C)} \lb \frac{R(T_0)R(T_0 \cap C)}{P(T_0)P(T_0 \cap C)} - \frac{R(T_0)^2}{P(T_0)^2} \rb \geq -\frac{\sqrt{\beta}\norm{w}_4^2}{P(C)}.
\end{align}   
\end{lemma}
\begin{proof}
If we have 
\[ \frac{R(T_0 \cap C)}{P(T_0 \cap C)} \geq \frac{R(T_0)}{P(T_0)} \]
then clearly both the LHSes are non-negative, hence both bounds hold.  
Assume this is not the case, then we have 
\begin{align*}
\frac{P(T_0 \cap C)}{P(C)} \lb \frac{R(T_0 \cap C)^2}{P(T_0 \cap C)^2} - \frac{R(T_0)R(T_0 \cap C)}{P(T_0)P(T_0 \cap C)}\rb &\geq 
- \frac{P(T_0 \cap C)}{P(C)} \frac{R(T_0)R(T_0 \cap C)}{P(T_0)P(T_0 \cap C)} \\
& \geq -\frac{P(T_0 \cap C)}{P(C)} \frac{R(T_0)^2}{P(T_0)^2} \\
& \geq -\frac{1}{P(C)} \frac{R(T_0)^2}{P(T_0)}
\end{align*}
and similarly
\begin{align}
\label{eq:before-gm}
 \frac{P(T_0 \cap C)}{P(C)} \lb \frac{R(T_0)R(T_0 \cap C)}{P(T_0)P(T_0 \cap C)} - \frac{R(T_0)^2}{P(T_0)^2}\rb & \geq -\frac{1}{P(C)} \frac{R(T_0)^2}{P(T_0)}.
\end{align}
We can bound this as
\begin{align*}
    \frac{R(T_0)^2}{P(T_0)} = P(T_0) \frac{R(T_0)^2}{P(T_0)^2} = \lb P(T_0) \cdot P(T_0) w(T_0)^4\rb ^{1/2} \leq \sqrt{\beta}\norm{w}_4^2
\end{align*}
    where we use $P(T_0) \leq \beta$ and $P(T_0)w(T_0)^4 \leq \norm{w}_4^4$. Plugging this into Equation \eqref{eq:before-gm} completes the proof.
\end{proof}

Finally for the set $T_1$ we show the following.
\begin{lemma}
\label{lem:t1-upper}
    For all $C \in \mC$, we have
\begin{align}
    \label{eq:t1-upper}
    \frac{P(T_1 \cap C)}{P(C)} \lb \frac{R(T_1 \cap C)^2}{P(T_1 \cap C)^2} - \frac{R(T_1)R(T_1 \cap C)}{P(T_1)P(T_1 \cap C)}\rb \geq  - \frac{\beta}{P(C)}, \\
    \label{eq:t1-lower}
    \frac{P(T_1 \cap C)}{P(C)} \lb \frac{R(T_1 \cap C)R(T_1)}{P(T_1 \cap C)P(T_1)} - \frac{R(T_1)^2}{P(T_1)^2}\rb \geq  - \frac{\beta}{P(C)}. 
\end{align}
\end{lemma}
\begin{proof}
If 
\[ \frac{R(T_1 \cap C)}{P(T_1 \cap C)} \geq \frac{R(T_1)}{P(T_1)} \]
then both LHSs are non-negative, so the bound holds. Else, 
\[ \frac{R(T_1 \cap C)}{P(T_1 \cap C)} \leq \frac{R(T_1)}{P(T_1)} \leq 1\] 
where the inequality is by second by the  definition of $T_1$. So we have the lower bound
\begin{align*}
\frac{P(T_1 \cap C)}{P(C)} \lb \frac{R(T_1 \cap C)^2}{P(T_1 \cap C)^2} - \frac{R(T_1)R(T_1 \cap C)}{P(T_1)P(T_1 \cap C)}\rb \geq -  \frac{P(T_1 \cap C)}{P(C)} \frac{R(T_1)R(T_1 \cap C)}{P(T_1)P(T_1 \cap C)} \geq -\frac{\beta}{P(C)}
\end{align*}
since $P(T_1 \cap C) \leq \beta$, and the other two ratios are at most $1$. This proves Equation \eqref{eq:t1-upper}. Equation \eqref{eq:t1-lower} is shown similarly.
\end{proof}

\begin{lemma}
\label{lem:ab-upper}
    For all $C \in \mC$, we have
    \begin{align*}
        \E_{\x \sim R|_C}[ w^*(\x)] + \frac{3}{R(C)}(\alpha \|w\|_2^2 +  \sqrt{\beta}\norm{w}_4^2) \geq \E_{\x \sim R|_C}[ w(\x)|].
    \end{align*}
\end{lemma}
\begin{proof}
 We first show the bound
    \begin{align}
    \label{eq:exp-p}
        \E_{\x \sim P|_C}[ w^*(\x)^2 - w(\x)w^*(\x)] \geq - \frac{3}{P(C)}(\alpha \|w\|_2^2 +  \sqrt{\beta}\norm{w}_4^2) .
    \end{align}
We have 
\begin{align}
\label{eq:last1}
 \E_{\x \sim P|_C}[ w^*(\x)^2] &= \E_{\s \sim P|_C} \E_{\x \sim P|_{\s \cap C}} [ w^*(\x)^2] \geq \E_{\s \sim P|_C} \left(\E_{\x \sim P|_{\s \cap C}} [ w^*(\x)]\right)^2\notag\\
&= \sum_{S \in \mS} \frac{P(S \cap C)}{P(C)}\frac{R(S \cap C)^2}{P(S \cap C)^2} \notag\\
&=  \sum_{i \in [m]}\frac{P(S_i \cap C)}{P(C)}\frac{R(S_i \cap C)^2}{P(S_i \cap C)^2} + \sum_{j \in \zo} \frac{P(T_j \cap C)}{P(C)} \frac{R(T_j \cap C)^2}{P(T_j \cap C)^2}.
\end{align}
On the other hand,
\begin{align}
    \E_{\x \sim P|_C}[w(x) w^*(\x)] &= \E_{\s \sim P|_C} \left[ w(\s)\E_{\x \sim P|_{\s \cap C}} [ w^*(\x)]\right] = \sum_{S \in \mS} \frac{P(S \cap C)}{P(C)}\frac{R(S)}{P(S)}\frac{R(S \cap C)}{P(S \cap C)} \notag\\
    &=  \sum_{i \in [m]}\frac{P(S_i \cap C)}{P(C)}\frac{R(S_i)R(S_i \cap C)}{P(S_i)P(S_i \cap C)} + \sum_{j \in \zo} \frac{P(T_j \cap C)}{P(C)} \frac{R(T_j)R(T_j \cap C)}{P(T_j)P(T_j \cap C)} \label{eq:last2}.
\end{align}
We subtract the Equation \eqref{eq:last2} from \eqref{eq:last1}. We then apply the lower bounds from Equation \eqref{eq:s-upper} to bound the contribution from the  $S_i$s, Equation \eqref{eq:t0-lower} for $T_0$ and Equation \eqref{eq:t1-lower} for $T_1$ to get
\begin{align*}
        \E_{\x \sim P|_C}[ w^*(\x)^2 - w(\x)w^*(\x)] &\geq - \frac{1}{P(C)}(3\alpha \norm{w}_2^2 + \sqrt{\beta}\norm{w}_4^2 + \beta)\\
        & \geq -\frac{3}{P(C)}(\alpha \norm{w}_2^2 + \sqrt{\beta}\norm{w}_4^2)
\end{align*}
which proves the bound claimed in Equation \eqref{eq:exp-p}.

 To derive the claim from this, we use the following equalities from Equation \eqref{eq:magic1} and \eqref{eq:magic2}:
\begin{align*}
    \E_{\x \sim P|_C}[w(\x)w^*(\x)] &= \E_{\x \sim R|_C}[ w(\x)]\frac{R(C)}{P(C)}, \\
    \E_{\x \sim P|_C}[w^*(\x)^2] &= \E_{\x \sim R|_C}[w^*(\x)]\frac{R(C)}{P(C)}.
\end{align*}
Plugging these into Equation \eqref{eq:exp-p} gives
\begin{align*}
    \frac{R(C)}{P(C)} \E_{\x \sim R|_C}[w^*(x) - w(x)] \geq  -\frac{3}{P(C)}(\alpha \norm{w}_2^2 + \sqrt{\beta}\norm{w}_4^2)
\end{align*}
which gives the claimed bound upon rearranging. 
\end{proof}

\begin{lemma}
\label{lem:ab-lower}
    For all $C \in \mC$, we have
    \begin{align*}
        \E_{\x \sim R|_C}[ w(\x)] + \frac{2}{R(C)}(\alpha \|w\|_2^2 +  \sqrt{\beta}\norm{w}_4^2) + \frac{2(\alpha + 2\beta)}{P(C)} \geq \E_{\x \sim P|_C}[ w^*(\x)|].
    \end{align*}
\end{lemma}
\begin{proof}
Recall that by Equation \eqref{eq:last2}
\begin{align*}
    \E_{\x \sim P|_C}[w(x) w^*(\x)] &=  \sum_{i \in [m]}\frac{P(S_i \cap C)}{P(C)}\frac{R(S_i)R(S_i \cap C)}{P(S_i)P(S_i \cap C)} + \sum_{j \in \zo} \frac{P(T_j \cap C)}{P(C)} \frac{R(T_j)R(T_j \cap C)}{P(T_j)P(T_j \cap C)}.
\end{align*}
Recall the bounds from Equations \eqref{eq:s-lower}, \eqref{eq:t0-lower} and \eqref{eq:t1-lower} which state
\begin{align*}
    \sum_{i \in [m]}\frac{P(S_i \cap C)}{P(C)}\lb \frac{R(S_i)R(S_i \cap C)}{P(S_i)P(S_i \cap C)} - \frac{R(S_i)^2}{P(S_i)^2}\rb  &\geq -\frac{\alpha \norm{w}_2^2}{P(C)}\\
    \frac{P(T_0 \cap C)}{P(C)} \lb \frac{R(T_0)R(T_0 \cap C)}{P(T_0)P(T_0 \cap C)} - \frac{R(T_0)^2}{P(T_0)^2} \rb &\geq -\frac{\sqrt{\beta}\norm{w}_4^2}{P(C)}\\
    \frac{P(T_1 \cap C)}{P(C)} \lb \frac{R(T_1 \cap C)R(T_1)}{P(T_1 \cap C)P(T_1)} - \frac{R(T_1)^2}{P(T_1)^2}\rb &\geq  - \frac{\beta}{P(C)}. 
\end{align*}
Adding these bounds, we get
\begin{align}
\label{eq:last3}
    \sum_{S \in \mS} \frac{P(S \cap C)}{P(C)} \frac{R(S)R(S \cap C)}{P(S)P(S \cap C)}  \geq \sum_{S \in \mS} \frac{P(S \cap C)}{P(C)} \frac{R(S)^2}{P(S)^2} -\frac{2}{P(C)}(\alpha \norm{w}_2^2 +  \sqrt{\beta}\norm{w}_4^2).
\end{align}

We also have 
\begin{align}
    \sum_{S \in \mS} \frac{P(S \cap C)}{P(C)}\frac{R(S)^2}{P(S)^2} &\geq  \lb\sum_{S \in \mS} \frac{P(S \cap C)}{P(C)}\frac{R(S)}{P(S)} \rb^2.\label{eq:last4} 
\end{align}
But note that 
\[ \sum_{S \in \mS} P(S \cap C)\frac{R(S)}{P(S)} = \sum_{S \in \mS} R(S)P(C|S) = Q(C) \geq R(C) - \alpha -2\beta \]
by Lemma \ref{lem:mc-ae} showing that $(\alpha, \beta)$-\mcab\ implies $(\alpha + 2\beta)$-\macc. Plugging this into Equation \eqref{eq:last4} gives
\begin{align}    
    \sum_{S \in \mS} \frac{P(S \cap C)}{P(C)}\frac{R(S)^2}{P(S)^2}  &\geq \lb\frac{R(C) - \alpha - 2\beta}{P(C)}\rb^2 \geq \lb\frac{R(C)}{P(C)}\rb^2 -2 (\alpha + 2\beta)\frac{R(C)}{P(C)^2}. \label{eq:last5}
\end{align}
Putting Equations \eqref{eq:last3} and \eqref{eq:last4} together with Equation \eqref{eq:last2} gives
\begin{align*}
    \E_{\x \sim P|_C}[w(\x)w^*(\x)] \geq \lb \E_{\s \sim P|C}[w^*(x)]\rb^2 - \frac{2}{P(C)}(\alpha \norm{w}_2^2 +  \sqrt{\beta}\norm{w}_4^2) - 2 (\alpha + 2\beta)\frac{R(C)}{P(C)^2} . 
\end{align*}
Using Equations \eqref{eq:magic1} and \eqref{eq:magic2} and diving both sides by $R(C)/P(C)$ gives 
\begin{align*}
    \E_{\x \sim R|_C}[w(\x)]  \geq \E_{\x \sim P|C}[w^*(\x)] - \frac{2}{R(C)}(\alpha \norm{w}_2^2 +  \sqrt{\beta}\norm{w}_4^2) - \frac{2(\alpha + 2\beta)}{P(C)} . 
\end{align*}
\end{proof}

\end{document}